\documentclass[journal]{IEEEtran}

\usepackage[ruled]{algorithm2e}
\usepackage{algorithmic}
\usepackage{color,xcolor,soul}
\usepackage{graphicx}
\usepackage{epstopdf}
\usepackage{mhchem}
\usepackage{subfigure}
\usepackage{bm}
\usepackage{tabularx}
\usepackage{colortbl,booktabs}
\usepackage{multirow}
\usepackage{adjustbox}
\usepackage{array}
\usepackage{makecell}
\usepackage[colorlinks,linkcolor=blue]{hyperref}

\newcommand{\PreserveBackslash}[1]{\let\temp=\\#1\let\\=\temp}
\newcolumntype{C}[1]{>{\PreserveBackslash\centering}p{#1}}
\newcolumntype{R}[1]{>{\PreserveBackslash\raggedleft}p{#1}}
\newcolumntype{L}[1]{>{\PreserveBackslash\raggedright}p{#1}}

\usepackage{array}
\usepackage{subfigure}
\graphicspath{{pic/}}

\begin{document}
	\title{A Unified Granular-ball Learning Model of Pawlak Rough Set and Neighborhood Rough Set}
	
	\author{Shuyin~Xia,
		Cheng~Wang,
		Guoyin~Wang*, \IEEEmembership{Senior~Member,~IEEE,}
		Weiping~Ding*,
		\IEEEmembership{Senior~Member,~IEEE,}
		Xinbo~Gao, 
		\IEEEmembership{Senior~Member,~IEEE,}
		Jianhang~Yu,
		Yujia~Zhai,
		Zizhong~Chen,  \IEEEmembership{Senior~Member,~IEEE}
		\IEEEcompsocitemizethanks{\IEEEcompsocthanksitem S. Xia, C. Wang, G. Wang, X. Gao \& J. Yu are with the Chongqing Key Laboratory of Computational Intelligence, Chongqing University of Telecommunications and Posts, 400065, Chongqing, China. E-mail: xiasy@cqupt.edu.cn, 1600029074@qq.com, wanggy@cqupt.edu.cn~(corresponding author), gaoxb@cqupt.edu.cn, yujianhang@cqupt.edu.cn.
			\IEEEcompsocthanksitem W. Ding is with the School of Information Science and Technology, Nantong University, Nantong 226019 China. E-mail: dwp9988@163.com~(corresponding author).
			\IEEEcompsocthanksitem Y. Zhai and Z. Chen are with the Department of Computer Science and Engineering, University of California at Riverside, Riverside, CA 92521, USA. E-mail: yzhai015@ucr.edu, chen@cs.ucr.edu.
	}}
	
	\markboth{arXiv}%
	{Shell \MakeLowercase{\textit{et al.}}: Bare Demo of IEEEtran.cls for Computer Society Journals}
	
	\markboth{arXiv}%
	{Shell \MakeLowercase{\textit{et al.}}: Bare Demo of IEEEtran.cls for IEEE Journals}
	
	\maketitle

	\begin{abstract}
Pawlak rough set and neighborhood rough set are the two most common rough set theoretical models. Pawlawk can use equivalence classes to represent knowledge, but it cannot process continuous data; neighborhood rough sets can process continuous data, but it loses the ability of using equivalence classes to represent knowledge. To this end, this paper presents a granular-ball rough set based on the granular-ball computing. The granular-ball rough set can simultaneously represent Pawlak rough sets, and the neighborhood rough set, so as to realize the unified representation of the two. This makes the granular-ball rough set not only can deal with continuous data, but also can use equivalence classes for knowledge representation. In addition, we propose an implementation algorithms of granular-ball rough sets. The experimental resuts on benchmark datasets demonstrate that, due to the combination of the robustness and adaptability of the granular-ball computing, the learning accuracy of the granular-ball rough set has been greatly improved compared with the Pawlak rough set and the traditional neighborhood rough set. The granular-ball rough set also outperforms nine popular or the state-of-the-art feature selection methods. All codes have been released in the open source GBRS library at http://www.cquptshuyinxia.com/GBRS.html.
	\end{abstract}
	

	\IEEEpeerreviewmaketitle

	 \vspace{-1em}	
       \section{Introduction}
        
        \IEEEPARstart{W}{ith} the advent of the era of big data, both the feasible size of datasets has grown explosively and the dimensionality of the data has increased dramatically\cite{01,65}. In machine learning and related fields, high dimensionality slows the training speeds of models and heightens the difficulty of the learning task. High dimensionality can lead to overfitting, which reduces the generalizability of a model \cite{02,03}. At high dimensions, Euclidean distance fails as a usable metric, limiting the application range of models that rely on Euclidean distance \cite{24,25,26,27,28}.   
    Feature engineering was developed in response to these substantial problems \cite{04,05}. Feature selection is one of the most well-known methods in feature engineering, which aims to select a subset of the feature set that can replace the original feature set. Feature selection has significant remedial effect: it can reduce feature dimension, increasing the training speed of a model; it can prevent overfitting, improving the generalizability of a model; and it can increase the correlation between features and predictions, making the model more interpretive\cite{66,67}. 
    
    Feature selection methods can be divided into three categories: filter methods, wrapper methods, and embedded methods. Filter methods score features according to evaluation criteria, then sort the features in descending order according to an assigned score. Evaluation criteria usually fall in one of four categories: distance measures, information measures, dependence measures, and consistency measures \cite{06}. Wrapper methods treat feature selection as a feature subset optimization process, and use a classifier to evaluate feature subsets. Since each feature subset needs to train a classifier, most wrapper methods are inefficient, and research into wrapper methods therefore usually focuses on the optimization process. Embedded methods embed feature selection into the training process of the learning algorithm to screen out which features are important for model training\cite{68,69}.
    
    Feature selection algorithms based on rough set theory rely on attribute reduction, which classes them as a filter method. Rough set theory was proposed by Polish scientist Zdzis\l{}aw Pawlak in 1982 \cite{07}. It is an effective mathematical tool to process uncertain, inconsistent, and incomplete data that has been widely applied in data mining, machine learning, decision support system, and other application fields \cite{08,09,10,11,12,13,14}. The classical rough set is also called Pawlak rough set. Pawlak rough set theory achieves this utility by using an equivalence relation to divide samples into several equivalence classes, and then defining upper and lower approximation sets using the union of the equivalence classes\cite{07}. The upper and lower approximations are used to describe and approximate uncertain concepts, and samples are divided into a positive region, boundary region and negative region during this process. The number of samples in the positive region is used to measure the dependence of a label on the feature set; that is, the score of the feature set. Heuristic attribute reduction algorithms based on rough set theory can effectively reduce the time complexity of high dimensional problems, making this a rich field of research in recent decades \cite{15,16,17,18,19,20,21}.
    
    Pawlak rough set (PRS) and neighborhood rough set (NRS) are two most popular rough set theories. In the feature selection process, PRS granulates a dataset based on equivalence classes. An equivalence class consists of a set of attributes and a set of objects, and can describe certain knowledge. So, it provides good interpretability. However, this also results in that it only can process discrete data. Data in the real world is mostly continuous data, the discretization of which will inevitably cause loss of information, thus presenting a serious hindrance to the development and application of rough set theory. To solve this problem, Hu, Yu, and Xie proposed the NRS model \cite{22} based on the idea of Lin's neighborhood model \cite{23}. NRS uses a neighborhood relation instead of an equivalence relation to granulate datasets, thus enabling NRS to process continuous data directly. However, in the model of NRS, the unpper and lower approximations of the NRS consist of sample points instead of equivalence classes, so the NRS loses interpretability. Besides, the inconsistency between PRS and NRS makes rough set theory not concise enough. 

    Our main contributions are as follows:
   
        \begin{enumerate}[\IEEEsetlabelwidth{4)}]
	\item We proposed an novel rough set model called granular-ball rough set for unifying PRS and NRS by introducing granular-ball into the rough set theory.
	
	\item The proposed granular-ball rough set is the first rough set model that can naturally process continuous data while having the interpretability of equivalence classes.
	
	\item Due to the combination of the robustness and adaptability of the granular-ball computing, the learning accuracy of the granular-ball rough set has been greatly improved compared with the PRS and NRS. The granular-ball rough set also outperforms other seven popular or the state-of-the-art feature selection methods. 
	
	\item As the GBRS can use equivalence class to represent upper and lower approximation while processing continuous data, we further proposed a granular-ball rough concept tree. This makes GBRS a strong mining tool that can process continuous data and realize feature selection, knowledge representation and classification at the same time. 
        \end{enumerate}
    
    The rest of this paper is organized as follows: we introduce related works in Section \ref{sec:relatedwork}. The theory basis of granular-ball rough set is presented in Section \ref{sec:theory}. Section \ref{sec:SDNRS} details our newly-proposed granular-ball rough set (GBRS) model, and experimental results and analysis are presented in Section \ref{sec:experiment}. We present our conclusion in Section \ref{sec:conclusion}.
    
    \section{Related Work\label{sec:relatedwork}}
    Rough set is mainly used for feature selection. So, in this section, we present a more detailed discussion of the prior work in the three categories of feature selection methods, as well as rough set theory. 
    \subsection{Filter Methods}
    The essence of filter methods is to use statistical indicators to score features, such as the Pearson correlation coefficient, the Gini coefficient, the Kullback-Leibler divergence, the Fisher score, similarity measures, and so forth. Since filter methods only use the dataset itself and do not rely on specific classifiers, they are very versatile and are easy to expand. When compared with wrapper and embedded methods, filter methods generally have a lower algorithm complexity. At the same time, the classification accuracy of filter methods is usually lowest among the three types of methods. Filter methods also only score a single feature, rather than an entire feature subset, and thus the feature subset generated by filter methods usually has high redundancy.
    
    Gu, Li, and Han proposed a generalized Fisher score feature selection method, aiming to find a feature subset that maximizes the lower bound of the Fisher score \cite{29}. This method transforms feature selection into quadratically-constrained linear program, and uses a cutting plane algorithm to solve the problem. Roffo and Melzi proposed a feature selection method based on graphs, ranking the most important features based on recognition as arbitrary clue sets \cite{30}. This method maps feature selection to an affinity graph by assigning features as nodes, and then evaluates the importance of each node via eigenvector centrality. In a later work, Roffo proposed the Inf-FS feature selection method, which assigns features as nodes of a graph and views feature subsets as paths in the graph \cite{31}. The power series property of matrices is used to evaluate the path, and the computational complexity is reduced by adding paths until the length reaches infinity.
    
    \subsection{Wrapper Methods}
    Wrapper methods use learning algorithms to evaluate features, and the classification accuracy of wrapper methods is often higher than that of filter methods. At the same time, the classifier used for evaluation limits the method, and the feature subset obtained by wrapper methods tends to have lower versatility.  For each feature subset, a wrapper method needs to train a classifier, resulting in a high computational complexity which depends on the search strategy of the feature subset. However, wrapper methods do evaluate the entire feature subset rather than a single feature and take into account the dependency between features, so the redundancy of the resulting feature subset is often lower than that of filter methods.
    
    The support vector machine (SVM) is a commonly used learning algorithm in wrapper methods. Guyon, Weston, Barnhill, and Vapnik proposed a feature selection method using SVM in combination with recursive feature elimination \cite{32}. The method constructs the ranking coefficient of features according to the weight vector generated by the SVM during training. In each iteration, the feature with the smallest ranking coefficient is removed, and finally a sort of all features in descending order is obtained. Guo, Kong, and He proposed a feature selection method based on clustering, which uses a triplet-based ordinal locality-preserving loss function to capture the local structures of the original data \cite{33}. The method defines an alternating optimization algorithm based on half-quadratic minimization to speed up the optimization process of this algorithm. Guo and Zhu specifically developed another wrapper method, Dependence Guided Unsupervised Feature Selection (DGUFS), to overcome the problem of single feature selection in filtering methods, using a joint learning framework for feature selection and clustering \cite{34}. DGUFS is a projection-free feature selection model based on $L_{2,0}$-norm equality constraints and two defined dependence-guided terms which increase the correlation between the original data, cluster labels, and the selected features.
    
    \subsection{Embedded Methods}
    Embedded methods embed feature selection into the learning algorithm, and the feature subset can be obtained when the training process of the learning algorithm has completed. This type of method is similar to filter methods, but the score of each feature is determined through model training. The idea behind these methods is to select those features important to the training of the model during the process of determining the model. Embedded methods are a compromise between filter methods and wrapper methods. Compared to filter methods, embedded methods can achieve a higher classification accuracy; compared to wrapper methods, embedded methods have lower algorithm complexity and are not as prone to overfitting.

    Bradley and Mangasarian proposed an embedded feature selection method based on concave minimization and SVM \cite{35}. This method finds a separation plane which distinguishes two point sets in the n-dimensional feature space while using as few features as possible. This method not only minimizes the weighted sum of the distance between incorrectly classified points and the boundary plane, but also maximizes the distance between the two boundary planes of the separation plane. Embedded methods are often based on regression learning algorithms. Nie, Huang, Cai, and Ding proposed an efficient and robust feature selection method using a loss function based on $L_{2,1}$-norms to remove outliers \cite{36}. This method adopts joint $L_{2,1}$-norm minimization on the loss function and regularization, and proposes an effective algorithm to solve joint $L_{2,1}$-norm minimization problems. Yang et al. proposed a  feature selection method, Unsupervised Discriminative Feature Selection (UDFS), which also uses the $L_{2,1}$-norm \cite{62}. UDFS optimizes an $L_{2,1}$-norm regularized minimization loss function, which uses discriminative information and the local structure of the data distribution.
    
    \subsection{Rough Set Theory}
    Feature selection methods based on rough set theory belong to the category of filtering methods. These methods use the positive region from rough set theory to score features. PRS granulates a dataset based on an equivalence relation, which provides good interpretability. his also result in that it only can process discrete data. However, data in the real world is mostly continuous data. Much research has been poured into overcoming the inability of rough set theory to process continuous data. This research can be roughly divided into two categories: discretizing continuous data or proposing improved rough set models. For decades, rough set models based on data discretization have proliferated \cite{37,38,39}. But the discretization of data will inevitably lead to the loss of information, and the discretization results will change with the discretization method. In light of this, some have proposed improved rough set models that can directly process continuous data. Dubois and Prade combine rough sets with another concept, fuzzy sets \cite{40}, and propose fuzzy rough sets \cite{41}, which replace the equivalence relation of classic rough sets with a fuzzy similarity relation, so that fuzzy rough sets can process continuous data. However, fuzzy rough set models need to set a membership function in advance using a priori knowledge of the dataset, which reduces the generality of fuzzy rough sets. 
    
    In contrast to fuzzy rough sets, NRS \cite{22} use a neighborhood relation to describe the relationships between samples. This neighborhood relation is completely derived from the data distribution and does not require any priori knowledge. At the same time, NRS can also process continuous data directly. Because of these advantages to NRS models, the field of NRS has been under continuous study and development. Li and Xie propose a method to accelerate NRS which based on an incremental attribute subset \cite{51}. Gao, Liu, and Ji use a matrix to preserve measurement calculation results, requiring only one dimension measurement calculation after a dimension increase and thereby reducing the amount of calculations require to find the positive region \cite{52}. In NRS, the neighborhood radius is a parameter that has a large impact on the reduction results and must be artificially set; how this parameter is chosen is also a frequent study of research. Peng, Liu, and Ji designed a fitness function, which combines the properties of datasets and classifiers to select the optimal neighborhood radius from a given neighborhood radius interval \cite{53}. Xia et al. propose an adaptive NRS model by combining granular ball computing with NRS, which can automatically optimize the neighborhood radius \cite{54}. Above NRS methods use a neighborhood relation instead of an equivalence relation to granulate datasets, thus enabling NRS to process continuous data directly. However, in the model of NRS, the unpper and lower approximations of the NRS consist of sample points instead of equivalence classes, so the NRS loses interpretability. Besides, the inconsistency between PRS and NRS makes rough set theory not concise enough. In this paper, We propose a novel rough set model named granular-ball rough set (GBRS) which can unify PRS and NRS. It not only has the interpretability of equivalence classes but can process continuous data naturally.
   
    \section{The Theory Basis of Granular-ball Rough Set \label{sec:theory}}
    In this section, in order to lay a foundation for our theorem and proof, we review some of the basic concepts of PRS and NRS, which have been presented in our previous work \cite{54}. In addition, granular-ball computing is the main basis of the proposed method, so we also introduce it in this section.
    
    \subsection{Pawlak Rough Set}
    We first introduce information system and indiscernible relation.
    
    \textbf{\emph{Definition 1.}} \cite{54} Let a quaternion $\left \langle U,A,V,f\right \rangle$ represent an \textit{information system} where:
    
    $ U=\left \{ x_{1},x_{2},...,x_{n} \right \}$ denotes a non-empty finite set of objects. $U$ is called the universe;
    
    $A=\left \{ a_{1},a_{2},...,a_{m} \right \}$ denotes a non-empty finite set of attributes;
    
    $V=\bigcup_{a\in A}V_{a}$ denotes the set of all attribute values, where $V_{a}$ denotes the value range of attribute $a$;
    
    $f=U\times A\rightarrow V$ denotes a mapping function: $\forall x_{i}\in U,a\in A $, $f\left ( x_{i},a \right )\in V_{a}$.
    
    This information system is called a \textit{decision system} $\left \langle U,C,D \right \rangle$ if the set of attributes in the information system above satisfies $A=C\cup D$, $C\cap D=\O$, and $D\neq \O $, where $C$ is the condition attribute set and $D$ is the decision attribute set.
    
    \textbf{\emph{Definition 2.}} \cite{54} Let $\left \langle U,A,V,f\right \rangle$ be an information system. $ \forall x,y\in U $ and $ B\subseteq A $, the \textit{indiscernible relation} $ IND(B) $ of the attribute subset $ B $ is defined as
    \begin{equation}
    IND(B)\!=\!\{(x, y) \in U \times U | f(x, a) = f(y, a), \forall a \in B\}.
    \end{equation}
    
    In PRS algorithms, $f(x,a) = V_a(x)$ represents $x$'s value on the attribute $a$. So, $f(x,a) = f(y,a)$ represents that the sample $x$ has the same value with the sample $y$ on the attribute $a$. In fact, $ (x, y) \in IND(B) $ shows that the values of samples $ x $ and $ y $ are the same under the attribute subset $ B $; that is, under the description of the attribute subset $ B $, samples $ x $ and $ y $ are indiscernible. 
    
    $ IND(B) $ is symmetric, reflexive, and transitive; that is, $ \forall B\subseteq A $, $ IND(B) $ is an equivalence relation on $ U $ (abbreviated as $ R_{B} $). $IND(B) $ creates a partition of $ U $, denoted $ U/IND(B) $ and abbreviated as $ U/B $. The characteristics of $ U/B $ are as follows: Suppose $ U/B=\{ X_{1},X_{2},...,X_{k} \} $, if $ X_{i}, X_{j}\subseteq U $, $ X_{i}\cap X_{j} = \O $($ i\neq j $), and $ \bigcup_{i=1}^{k}X_{i}=U $, then $ U $ is divided into $ k $ parts by $ IND(B) $. An element $ [x]_{B}=\{y \in U |(x, y) \in IND(B)\} $ in $ U/B $ is called an equivalence class. This leads us to our next set of definitions, approximations based on the equivalence relation $R_B$.
    
    \textbf{\emph{Definition 3.}} \cite{54} Let $\left \langle U,A,V,f\right \rangle $ be an information system. $ \forall B \subseteq A $, there is a corresponding equivalence relation $ R_{B} $ on $ U $. Then, $ \forall X \subseteq U $, the \textit{upper and lower approximation of} $ X $ with respect to $ B $ are defined as follows:
    \begin{equation}
    \overline{R_{B}}X=\cup\left\{[x]_{B} \in U / B |[x]_{B} \cap X \neq \emptyset\right\},
    \end{equation}
    \begin{equation}
    \underline{R_{B}}X=\cup\left\{[x]_{B} \in U / B |[x]_{B} \subseteq X\right\}.
    \end{equation}
    The lower approximation $ \underline{R_{B}}X $ represents the set of samples in $ U $ that are determined to belong to $ X $ according to the equivalence relation $ R_{B} $. It essentially reflects the ability of the equivalence relation $ R_{B} $ to approximately describe the knowledge contained in $ X $ by a partition of the knowledge of the universe $ U $. It is also commonly called the \textit{$ B $ positive region of $ X $ in $ U $}, which is abbreviated as $ POS_{B}(X) $.
    
    \textbf{\emph{Definition 4.}} \cite{54} Let $ \left \langle U,C,D \right \rangle $ be a decision system. We notate the partition of the universe $U$ by the decision attribute set $D$ into $L$ equivalence classes by $ U/D=\left \{ X_{1},X_{2},...,X_{L} \right \}$. $ \forall B\subseteq C$, there is a corresponding equivalence relation $ R_{B} $ on $ U $. The \textit{upper} and the \textit{lower approximation of $D$ with respect to $B$} are respectively defined as
    \begin{equation}
    \overline{R_{B}}D=\bigcup_{i=1}^{L}\overline{R_{B}}X_{i},
    \end{equation}
    \begin{equation}
    \underline{R_{B}}D= \bigcup_{i=1}^{L}\underline{R_{B}}X_{i}.
    \end{equation}
    
    \textbf{\emph{Definition 5.}} \cite{54} Let $ \left \langle U,C,D \right \rangle$ be a decision system. $\forall B\subseteq C$, the \textit{positive region} and \textit{boundary region} of $D$ with respect to $B$ are respectively defined as:
    \begin{equation}
    POS_{B}\left ( D \right )=\underline{R_{B}}D,
    \end{equation}
    \begin{equation}
    BN_{B}\left ( D \right )=\overline{R_{B}}D-\underline{R_{B}}D.
    \end{equation}
    
    The size of the positive region reflects the separability of the classification problem in a given attribute space. The larger the positive region, the more detailed the classification problem can be described using this attribute set. We find it useful to describe this mathematically: the \textit{dependence} of $D$ on $B$ is defined as
    \begin{equation} \label{equ:dependence}
    \gamma _{B}\left ( D \right )=\frac{\left | POS_{B}\left ( D \right ) \right |}{\left | U \right |},
    \end{equation}
    where $\left | \cdot  \right |$ is the cardinality of the set and $0\leq \gamma _{B}\left ( D \right )\leq 1$. Obviously, the larger the positive region, the stronger the dependence of $D$ on $B$.
    
    The dependency function defines the contribution of conditional attributes to a classification, so it can be used as an evaluation index for the importance of the attribute set.
    
    \textbf{\emph{Definition 6.}} \cite{54} Given a decision system $ \left \langle U,C,D \right \rangle$, $\forall B\subseteq C$ and $\forall a\in(C-B) $, the \textit{importance} of $a$ relative to $B$ is defined as
    \begin{equation}
    SIG\left ( a,B,D \right )=\gamma _{B\cup a}\left ( D \right )- \gamma _{B}\left ( D \right ).
    \end{equation}
    
    Rough set uses the measurement $SIG$ in (9) to select attributes in a forward way. The selection result $C'$ is initialized with $\O$, and for each attribute $a$ in the attribute $C-C'$, that with the largest value of $SIG(a,C', D)$ which should be larger than 0 is select into $C'$. This process is repeated until all $SIG(a,C', D)$ is not greater than 0.

    \subsection{Neighborhood Rough Set}
    After introducing NRS somewhat loosely, we now drill down into the details, defining the basic spaces we are operating in, the neighborhoods we are working with in NRS, and the positive region we have mentioned, which is key to the operation of these methods.

    \textbf{\emph{Definition 7.}} \cite{54} Let $\Delta : \Omega\times \Omega\rightarrow R$ be a function generated on a set $\Omega$. $\left \langle \Omega ,\Delta  \right \rangle$ is known as a \textit{metric space} if $\Delta$ satisfies:
    
    (1) $\Delta \left ( x_{1},x_{2} \right )\geq 0$, $\Delta \left ( x_{1},x_{2} \right )=0$ if $x_{1}$=$x_{2},\forall x_{1},x_{2}\in\Omega$;
    
    (2) $\Delta \left ( x_{1},x_{2} \right )=\Delta \left ( x_{2},x_{1} \right ),\forall x_{1},x_{2}\in\Omega$;
    
    (3) $\Delta \left ( x_{1},x_{3} \right )\leq  \Delta \left ( x_{1},x_{2} \right )+ \Delta \left ( x_{2},x_{3} \right ),\forall x_{1},x_{2},x_{3}\in\Omega$.
    
    \noindent In this case, $\Delta$ is known as a \textit{metric}.    
    
    \textbf{\emph{Definition 8.}} \cite{54} Let $U=\left \{ x_{1},x_{2},...,x_{n} \right \}$ be a non-empty finite set of real space. $\forall x_{i}\in U $, the $\delta$\textit{-neigborhood} of $x_i$ is defined as:
    \begin{equation}
    \delta \left ( x_{i} \right )=\left \{ x\mid x\in U,\Delta \left ( x,x_{i}  \right )\leq \delta \right \},
    \end{equation}
    where $\delta \geq 0 $.
    
    \textbf{\emph{Definition 9.}} \cite{54} Let $\left \langle U,C,D \right \rangle$ be a neighborhood decision system. The decision attribute set $D$ divides $U$ into $L$ equivalence classes: $ X_{1},X_{2},...,X_{L}$. $\forall B\subseteq C$, the \textit{lower approximation} and the \textit{upper approximation} of the decision attribute set $D$ with respect to the condition attribute set $B$ are respectively defined as:
    \begin{equation}
    \underline{N_{B}}D= \bigcup_{i=1}^{L}\underline{N_{B}}X_{i},
    \end{equation}
    \begin{equation}
    \overline{N_{B}}D=\bigcup_{i=1}^{L}\overline{N_{B}}X_{i},
    \end{equation}
    where $ \underline{N_{B}}X_{i}=\left \{ x_{k}\mid \delta _{B}\left ( x_{k} \right )\subseteq X_{i},x_{k}\in U\right \}$, $\overline{N_{B}}X_{i}=\left \{ x_{k}\mid \delta _{B}\left ( x_{k} \right )\bigcap X_{i}\neq \O ,x_{k}\in U\right \}$, and its \textit{positive region} and \textit{boundary region} are respectively defined as $ POS_{B}\left ( D \right )=\underline{N_{B}}D,BN\left ( D \right )=\overline{N_{B}}D-\underline{N_{B}}D$.
    
    \subsection{Granular-ball Computing}
    
\begin{figure}
    	\centering
    	\includegraphics[scale=0.15]{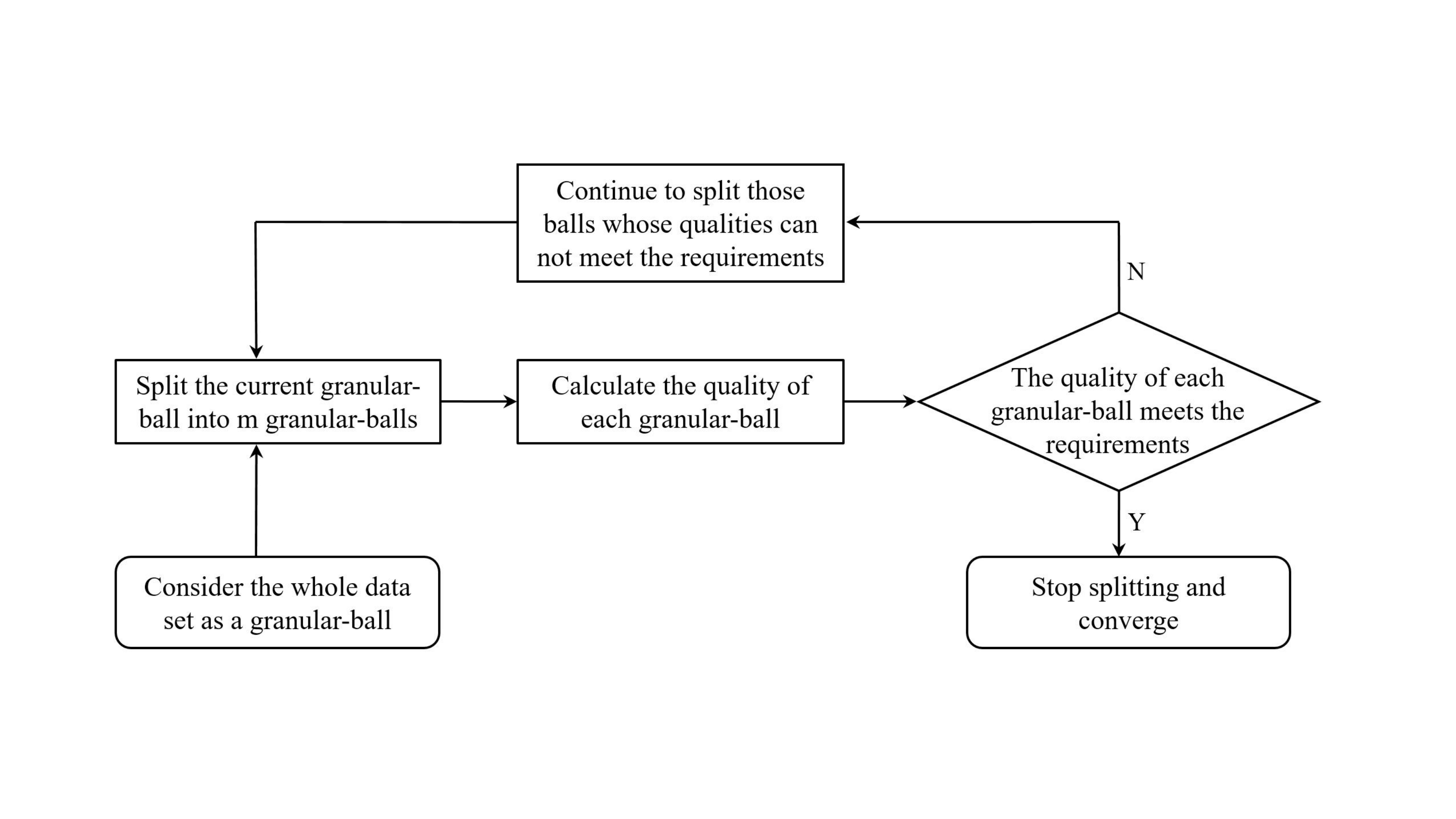}
    	\caption{Process of granular-ball generation in granular-ball computing.} 
    	\label{fig:GBCProcess}	
    \end{figure}

    Combining the theoretical basis of traditional granular computing, and based on the research results published by Chen in Science in 1982, he pointed out that ``human cognition has the characteristics of large-scale priority" {\cite{55}}, Wang put forward a lot of granular cognitive computing{\cite{56}}. Based on granular cognitive computing, granular-ball computing is a new, efficient and robust granular computing method proposed by Xia and Wang \cite{57}, the core idea of which is to use ``granular-balls'' to cover or partially cover the sample space. A granular-ball $GB=\{x_i,i=1...N\}$, where $x_i$ represents the objects in $GB$, and $N$ is the number of objects in $GB$. $GB$'s center $C$ and radius $r$ are respectively represented as follows
    \begin{equation}
     C=\frac {1}{N}{}\sum\limits_{i=1}^{N}{x_{i}},
    \end{equation}
    \begin{equation}
    r=\frac{1}{N}\sum\limits_{i=1}^{N}{\left| {{x}_{i}}-C \right|}.
    \end{equation}
    This means that the radius is equal to the average distance from all objects in $GB$ to its center. The radius can also be set to the maximum distance. The ``granular-ball'' with a center and radius are used as the input of the learning method or as accurate measurements to represent the sample space, achieving multigranularity learning characteristics (that is, scalability, multiple scales, etc.) and the accurate characterization of the sample space. The basic process of granular-ball generation for classification problems in granular-ball computing is shown in Figure \ref{fig:GBCProcess}.
    
    As shown in Figure \ref{fig:GBCProcess}, to similiate the ``the characteristics of large-scale priority of human cognition'' at the beginning of the algorithm, the whole dataset can be regarded as a granular-ball. At this time, the purity of the granular-ball is the worst and cannot describe any distribution characteristics of the data. The ``purity'' is used to measure the quality of a granular-ball \cite{57}. It is equal to the proportion of the most labels in the granular-ball. Then, the number of labels $m$ of different classes in the granular-ball is counted, and the granular-ball can be split into $m$ granular-balls. The next step is to calculate the purity of each granular-ball. This is the key step, because purity is the criterion for evaluating whether a granular-ball needs to continue to split. As the splitting process continues to advance, the purity of the granular-balls increases, and the decision boundary becomes increasingly clearer; until the purity of all granular-balls meets the requirements, the boundary is clearest, and the algorithm converges. The granular-ball computing has developed granular-ball classifiers \cite{57}, granular-ball clustering \cite{58}, granular-ball neighborhood rough set \cite{54} and granular-ball sampling methods \cite{59}.

    \section{Granular-ball Rough Sets}\label{sec:SDNRS}
    \subsection{Motivation}
\begin{figure}[!ht]
    	\centering
    	{\includegraphics[width = 0.4\textwidth]{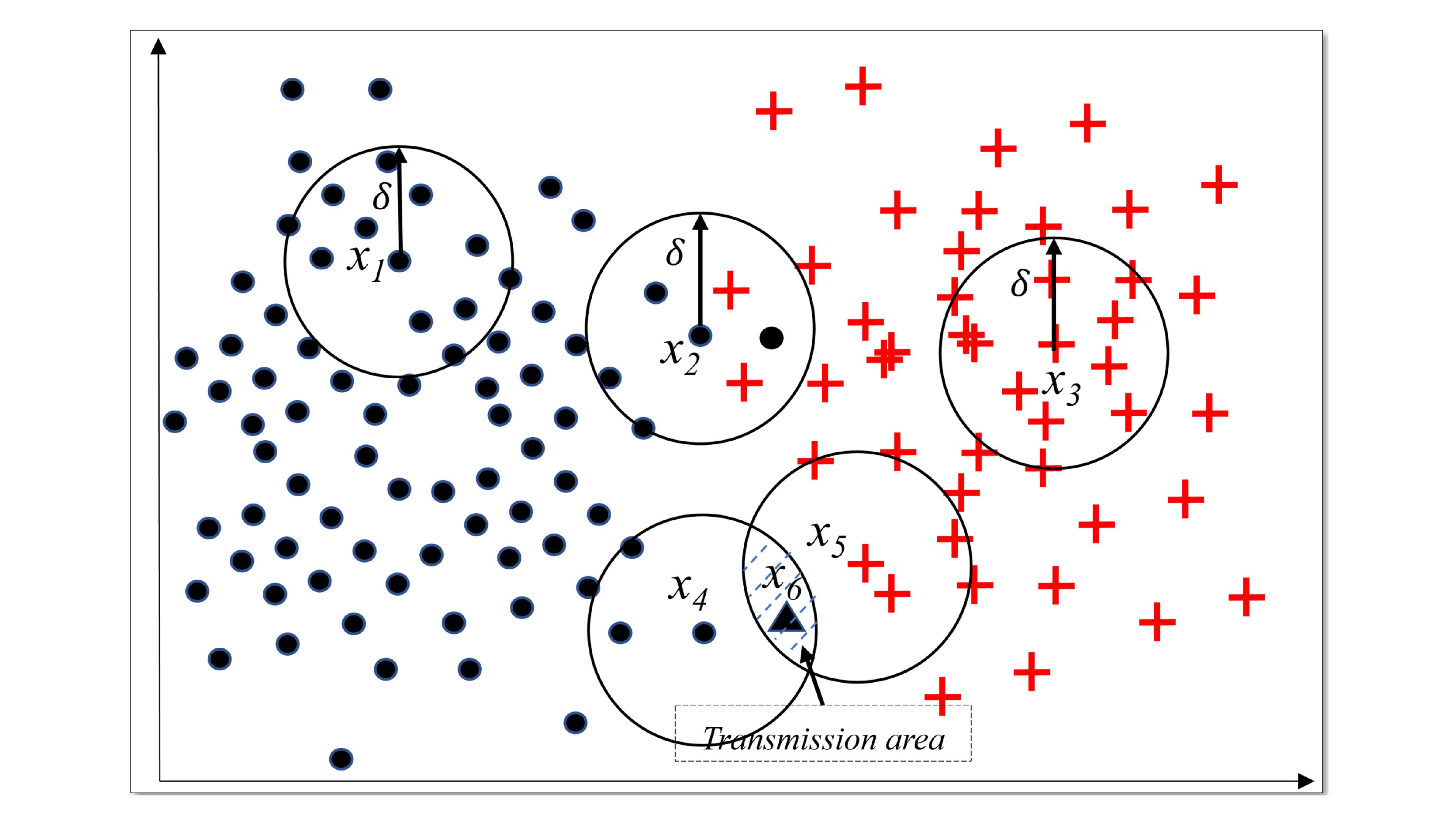}}
    	\caption{The phenomena of the heterogeneous transmission. The label of a black circle point is equal to ``+1'', and the label of a red plus point is equal to ``-1''. The triangle point $x_6$ is a new test sample.}
    	\label{fig:HeterTrans}
    \end{figure}

    The main difference of upper or lower approximation between the model of PRS and NRS is that, as shown in Definition 3 and Definition 10 respectively, the former consists of equivalence classes, which can be used to represent knowledge and has the interpretability; however, the latter consists of sample points, which has no interpretability. If we want to use equivalence classes to describe the upper and lower approximation of NRS, a straightforward approach is to treat all objects in a neighborhood radius as an equivalence class. However, we find that this may make two equivalence classes with different decision labels equal. We called this phenomen as ``heterogeneous transmission''. It can be described in detail in Figure \ref{fig:HeterTrans}. As shown in Figure \ref{fig:HeterTrans}, according with the Definition 10, those objects including $x_1, x_3, x_4, x_5$ belong to positive region, and $x_2$ belongs to boundary region. The heterogeneous transmission appears in the intersecting area of the neighborhood area of $x_4$ and that of $x_5$. The intersecting area is called ``transmission area''. When we define the objects belong to a given neighborhood as a equivalence class, the label of the neighborhood equivalence class of $x_4$ is equal to ``+1'', and the label of the neighborhood equivalence class of $x_5$ is equal to ``-1''. However, a new object $x_6$ in transmission area is equivalent to the neighborhood equivalence class of $x_4$ and that of $x_5$ at the same time. This makes the two equivalence classes with two different label equivalent. Obviously, it is harmful for learning. To avoid the heterogeneous transmission phenomena, a method is to set the neighborhood radius small enough. However, this may make most of objects or all objects always belong to positive region, and positive region can not be effectively used for measuring feature importance or other learning tasks. Overall, the heterogeneous transmission phenomena is caused by the overlap between those positive region neighborhoods with different labels in NRS. 
    
    \begin{figure}[!ht]
	\centering
	{\includegraphics[width = 0.4\textwidth]{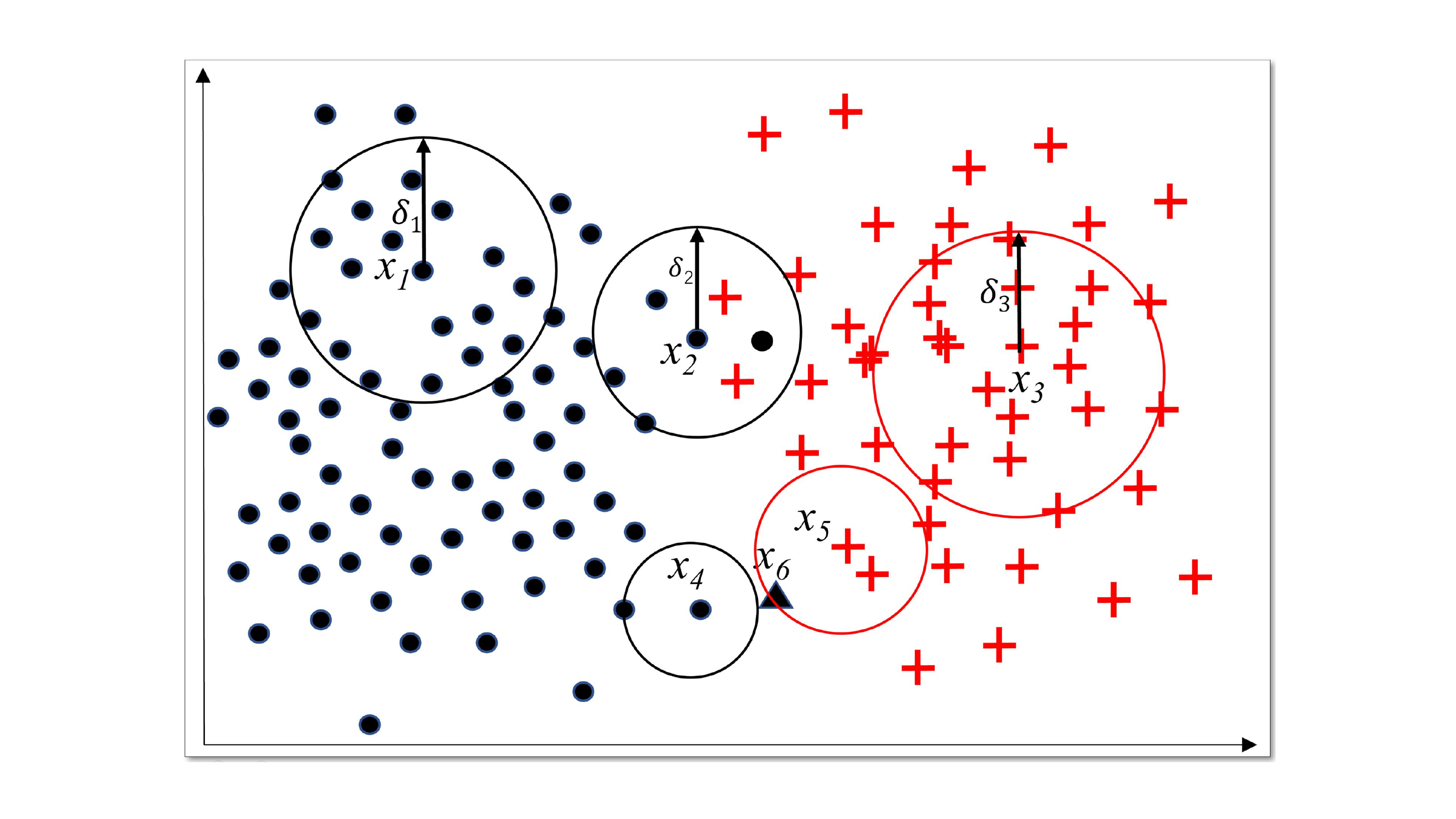}}
	\caption{The phenomena of the heterogeneous transmission in Fig. 2 is removed using granular-ball computing.}
	\label{fig:NoHeterTrans}
\end{figure}
    
    \subsection{Granular-ball Rough Set}
    As shown in Fig. \ref{fig:NoHeterTrans}, as the neighborhood radii are adaptively different, the overlap between those positive region neighborhoods does not exist in granular-ball computing. Therefore, it is possible to use equivalence classes to represent upper and lower approximation by introducing granular-ball computing to represent neighborhood. The granular-ball computing based rough set is called ``granular-ball rough set'', and its models are defined and described as follows.

    \textbf{\emph{Definition 10.}} Let $U=\left \{ x_{1},x_{2},...,x_{n} \right \}$ be a non-empty finite set of real space. $\forall x_{i}\in U $, a granular-ball $GB_j$ is defined as:
    \begin{equation}
    GB_j =\left \{ x\mid x\in U,\Delta \left ( x,c_{j}  \right )\leq r _j, \right \},
    \end{equation}
    where $c_j$ and $r_j$ denote the center and the radius of $GB_j$ respectively. The larger the $r_i$, the coarser the granular-ball $GB_j$; otherwise, the finer the granular-ball $GB_j$.
    
    \textbf{\emph{Definition 11.} \label{def:indgb}}  Let $\left \langle U,A,V,f\right \rangle$ be an information system. $ \forall x,y\in U $ and $ B\subseteq A $, the \textit{indiscernible granular-ball relation} $ INDGB(B) $ of the attribute subset $ B $ is defined as
    \begin{equation}
    INDGB(B)\!=\!\{(x, y) \in U^{2} | f(x, a) = f(y, a) = GB , \forall a \in B\}.
    \end{equation}
    If $(x,y) \in INDGB(B)$, the relationship between $x$ and $y$ is denoted as $x \sim y$.
    
    In granular-ball rough set, $f(x,a) = GB, x \in GB$. So, $f(x, a) = f(y, a) = GB$ represents that $x$ and $y$ belong to the same granular-ball under the given attribute set $a$. $ \forall B\subseteq A $, $ INDGB(B) $ is an equivalence relation on $ U $ (abbreviated as $ GBR_{B} $). Because the granular-balls do not overlap, $ GBIND(B) $ can also create a partition of $ U $, denoted $ U/GBIND(B) $ and abbreviated as $ U/GB(B) $. An element $ [x]_{GB(B)}=\{y \in U |(x, y) \in INDGB(B)\} $ in $ U/GB(B) $ is an equivalence class generated by granular-ball computing.

    \textbf{\emph{Definition 12.}} \label{def:GBE} Given an information system $\left \langle U,A,V,f\right \rangle$, if $GB_i \cup GB_j \neq \emptyset$, $GB_i \sim GB_j$.
    
    As the overlap does not exist between those granular-balls with different labels in GBRS, Definition 12 means that those granular- balls with the same label belong to an equivalence class. This kind of overlap between those positive region neighborhoods, i.e., granular-balls, with a same label is not considered in this method because it does not lead to heterogeneous transmission and affect decision; besides, considering this overlap in the algorithm design will increase computation cost.
    
    \textbf{\emph{Properties of GBRS.}}  Given an information system $\left \langle U,A,V,f\right \rangle$, $x,y,z \in U$, $ B\subseteq A $, $\sim$ represents the indiscernible granular-ball relation of the attribute subset $B$ on $U$. The indiscernible granular-ball relation obviously has the following properties:
    \begin{enumerate}
    	\item[(1)] Symmetry: if $x\sim y$, then $y\sim x$;
    	\item[(2)] Reflexivity: $x\sim x$ ;
    	\item[(3)] Transitivity: if $x\sim y$, $y\sim z$, then $x\sim z$.	
    \end{enumerate}

    In summary, similar with that in PRS, $ INDGB(B) $ is symmetric, reflexive, and transitive, and complete consistent with $IND(B)$ in PRS. 
    
    Based on the equivalence class $[x]_{GB(B)}$, the definitions of positive region, upper and lower approximations are the same to those in PRS. Therefore, GBRS has the consistent model with the PRS. Their specific definitions are as follows:
    
    \textbf{\emph{Definition 13.}} Let $\left \langle U,A,V,f\right \rangle $ be an information system. $ \forall B \subseteq A $, there is a corresponding equivalence relation $ GBR_{B} $ on $ U $. Then, $ \forall X \subseteq U $, the \textit{upper and lower approximation of} $ X $ with respect to $ B $ are defined as follows:
    \begin{equation}
    \overline{GBR_{B}}X=\cup\left\{[x]_{B} \in U / GB(B) |[x]_{GB(B)} \cap X \neq \emptyset\right\},
    \end{equation}
    \begin{equation}
    \underline{GBR_{B}}X=\cup\left\{[x]_{B} \in U / GB(B) |[x]_{GB(B)} \subseteq X\right\}.
    \end{equation}
    
    \textbf{\emph{Definition 14.}} Let $ \left \langle U,C,D \right \rangle $ be a decision system. We notate the partition of the universe $U$ by the decision attribute set $D$ into $L$ equivalence classes by $ U/D=\left \{ X_{1},X_{2},...,X_{L} \right \}$. $ \forall B\subseteq C$, there is a corresponding equivalence relation $ GBR_{B} $ on $ U $. The \textit{upper} and the \textit{lower approximation of $D$ with respect to $B$} are respectively defined as
    \begin{equation}
    \overline{GBR_{B}}D=\bigcup_{i=1}^{L}\overline{GBR_{B}}X_{i},
    \end{equation}
    \begin{equation}
    \underline{GBR_{B}}D= \bigcup_{i=1}^{L}\underline{GBR_{B}}X_{i}.
    \end{equation}
    
    According to Definition 14, a granular-ball whose purity is equal to 1, i.e., that the samples in it have a same decision label, belongs to lower approximation (i.e., the positive region described in Definition 15) in a decision system.
    
    \textbf{\emph{Definition 15.}} Let $ \left \langle U,C,D \right \rangle$ be a decision system. $\forall B\subseteq C$, the \textit{positive region} and \textit{boundary region} of $D$ with respect to $B$ are respectively defined as:
    \begin{equation}
    POS_{B}\left ( D \right )=\underline{GBR_{B}}D,
    \end{equation}
    \begin{equation}
    BN_{B}\left ( D \right )=\overline{GBR_{B}}D-\underline{GBR_{B}}D.
    \end{equation}
    
    Completely the same with that in the Pawlak rough set, the size of the positive region reflects the separability of the classification problem in a given attribute space. The larger the positive region, the more detailed the classification problem can be described using this attribute set. The \textit{dependence} of $D$ on $B$ and $SIG\left ( a,B,D \right )$ are also the same with that in Pawlak rough set.    
    
    In the models of GBRS, when the radius of each granular-ball is set to a infinitely small positive number, GBRS is transformed into PRS. When the PRS algorithm is designed from the perspective of GBRS, it is called as granular-ball PRS (GBPRS). GBPRS and PRS have the same experimental results; however, their algorithm designs are different, and the former generate eqivalence using granular-ball computing. When the radius of each granular-ball is not set to a zero, GBRS is transformed into granular-ball NRS (GBNRS). As the GBNRS not only can use equivalence classes to represent knowledge, but is much more efficient than the traditional NRS which contains many overlaps, the GBNRS can completely replace the traditional NRS. In another word, the GBNRS is the representive method of neighborhood rough set. In summary, GBRS is an unified model of GBPRS and GBNRS. 
    
    In addition, as shown in Fig. 5(f), GBNRS can flexibly fit the data distribution using those granular-balls with varies radii, which is obviously better than those methods using a fixed radius, such as PRS and the traditional NRS. So, GBNRS can achieve a higher accuracy than the two algorithms. Moreover, the combination of the robustness and adaptability of the granular-ball computing is helpful for GBNRS to perform well in accuracy. This robustness in the GBNRS is reflected in the fact that, since the noise point will be in the small granular-ball, the characteristics of a large neighborhood, i.e., whether it belongs to positive region or not, will not be affected by it. This robustness will not exist in other most methods, such as the traditional NRS who has a fixed radius. These will be demonstrated in the experiments.
    
    \subsection{Implement of GBNRS}

\begin{figure}[!ht]
    	\centering
    	{\includegraphics[scale=0.26]{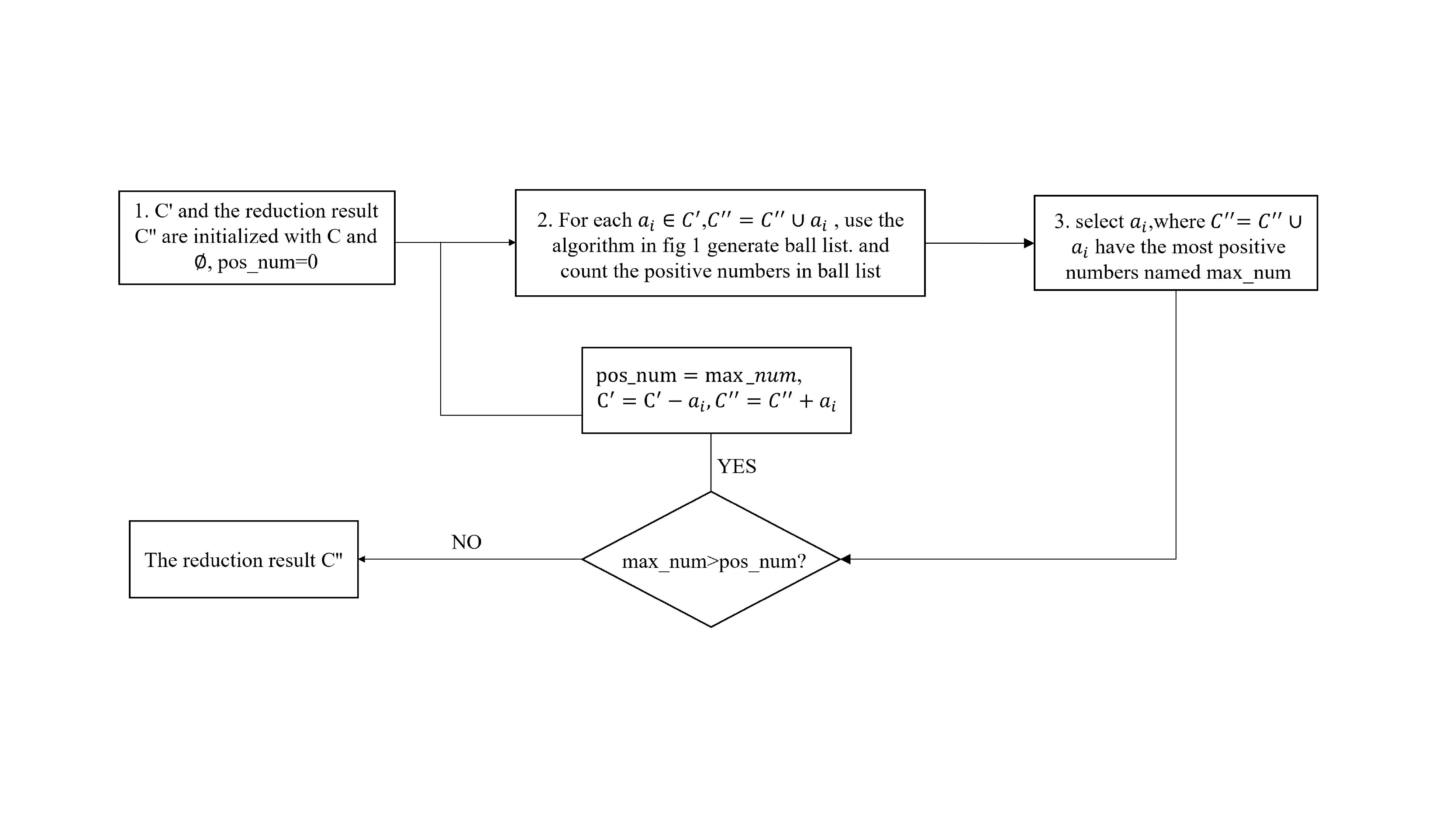}}
    	\caption{Process of attribute reduction in GBRS.}
    	\label{fig:reductionprocess}
      \end{figure}

   As the GBNRS has the unified model with the PRS, as shown in Fig. \ref{fig:reductionprocess}, its whole algorithm process is completely the same with that of the PRS. The only difference bewteen the GBRS and PRS is the generation way of positive region, which is shown in step 2 in Fig. \ref{fig:reductionprocess}. The GBRS generates positive region using granular-balls. For the GBNRS, in the granular-ball generation, to fulfilling definition 14, the purity threshold $PT$ is set to 1. Besides, referring to that in \cite{70}, the lower bound of the size of a granular ball $LBS$, i.e., the number of samples in it, is optimized from 2*$d$ to 2 with a step as 1, where the $d$ denotes the number of conditional attributes in the data set. When the size of a granular-ball is lower than $LBS$ or its purity reaches to 1, the granular ball stop to split. According to Definition 15, a granular-ball whose purity is equal to 1 belongs to positive region, and a granular-ball whose purity is lower than 1 belongs to boundary region. Besides, to decrease the randomness in the granular-ball generation and make the positive region of the granular-balls in the attribute selection process have better comparability, for a granular-ball, the samples who have the smallest indexes in it are selected as the initial centroids in this granular-ball splitting process. The flowchart is shown as Fig. \ref{fig:reductionprocess}. The specific process is mainly composed of five steps.
    
    \begin{figure}[!t]      
    	\centering
    	\subfigure[]{\includegraphics[width=0.24\textwidth]{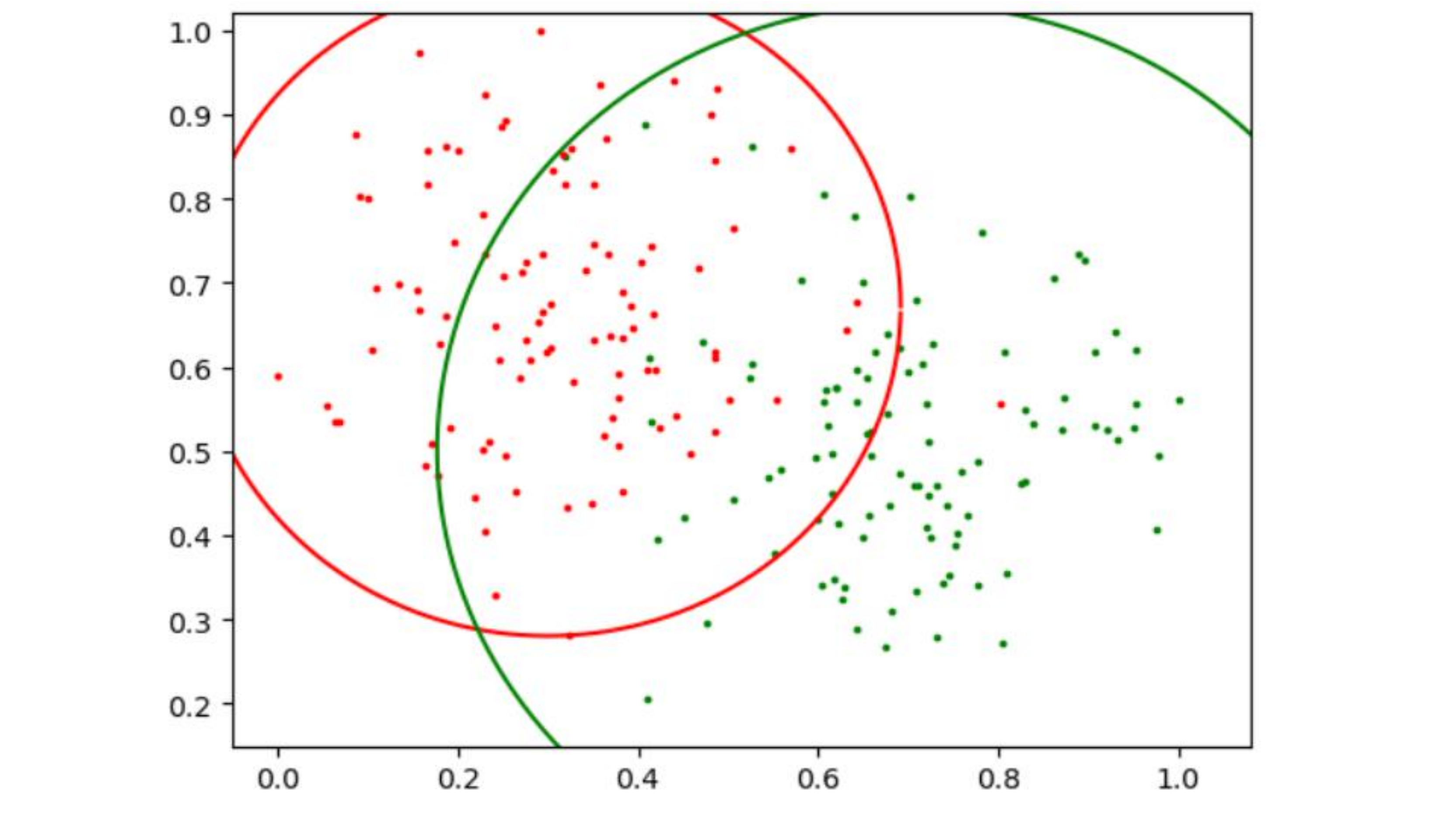}}
    	\subfigure[]{\includegraphics[width=0.24\textwidth]{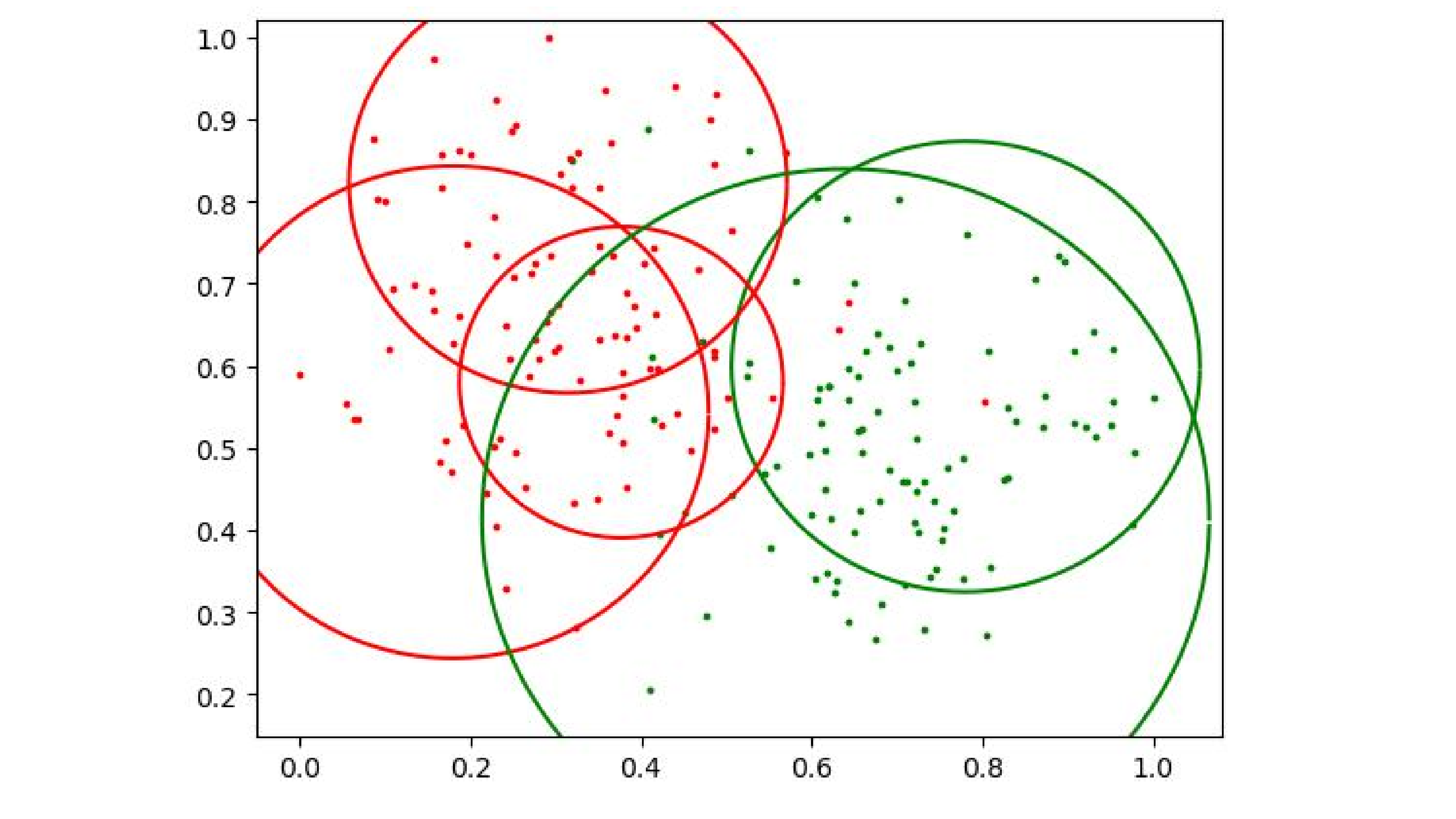}}
    	\subfigure[]{\includegraphics[width=0.24\textwidth]{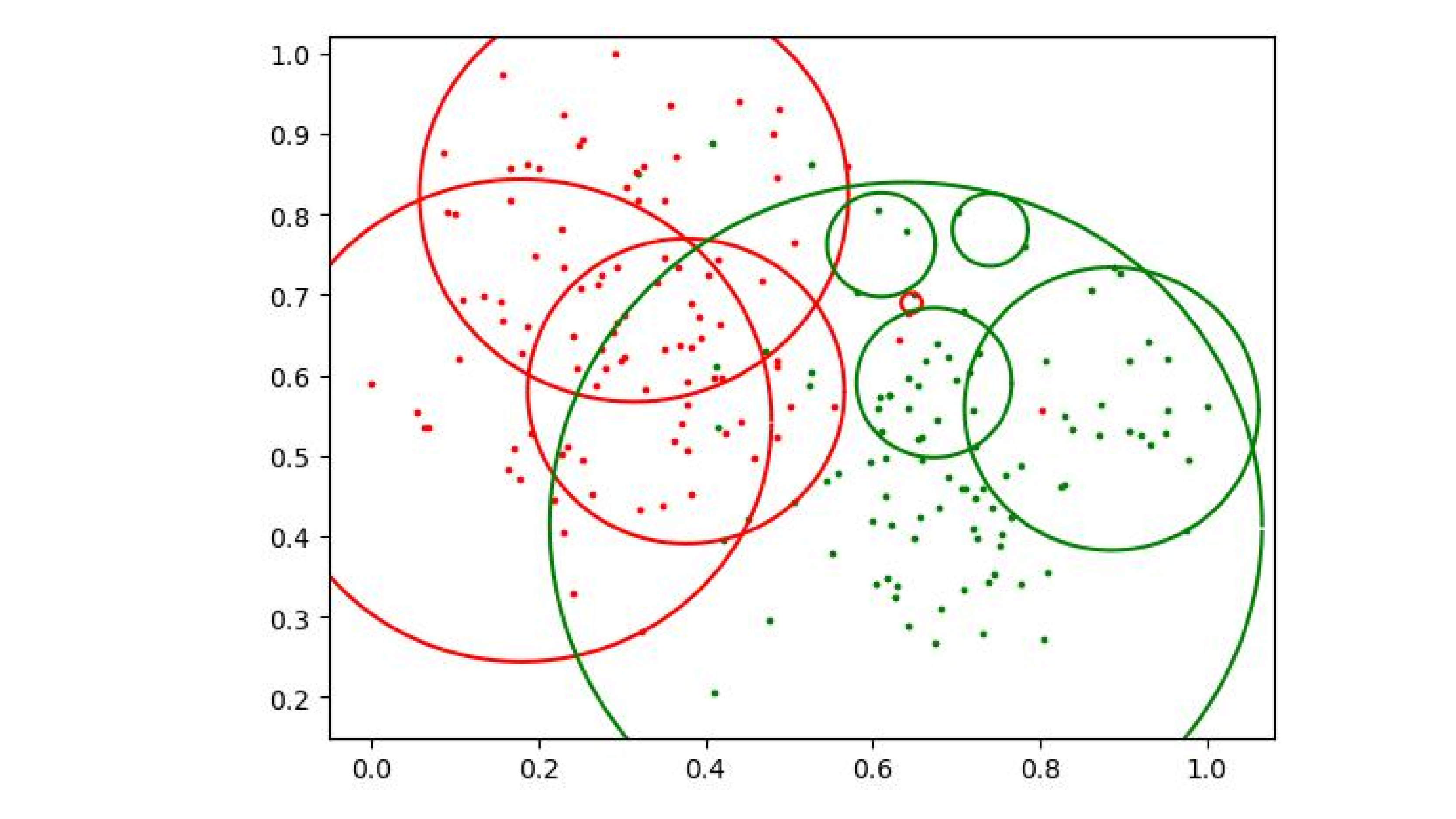}}
    	\subfigure[]{\includegraphics[width=0.24\textwidth]{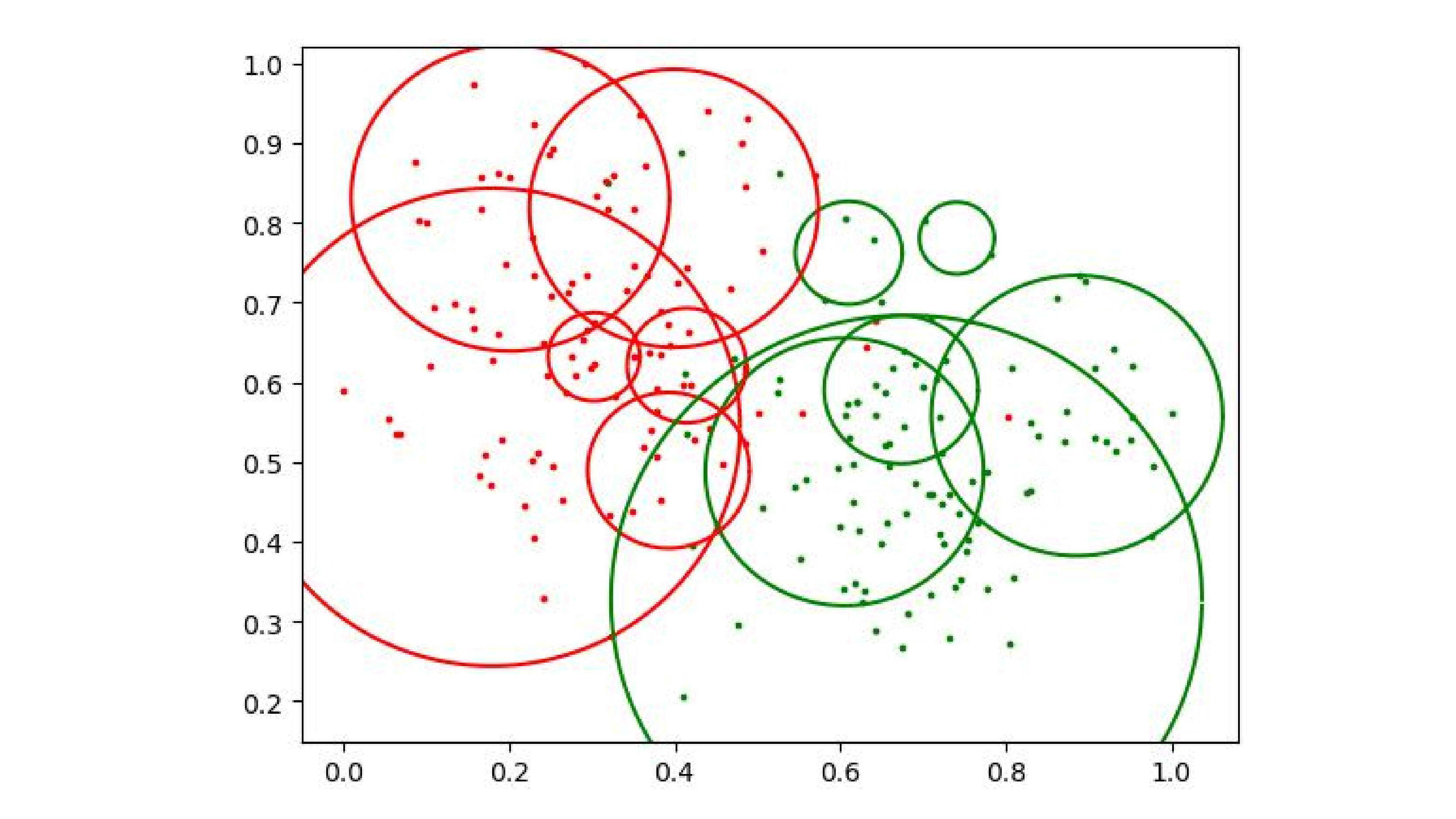}}
    	\subfigure[]{\includegraphics[width=0.24\textwidth]{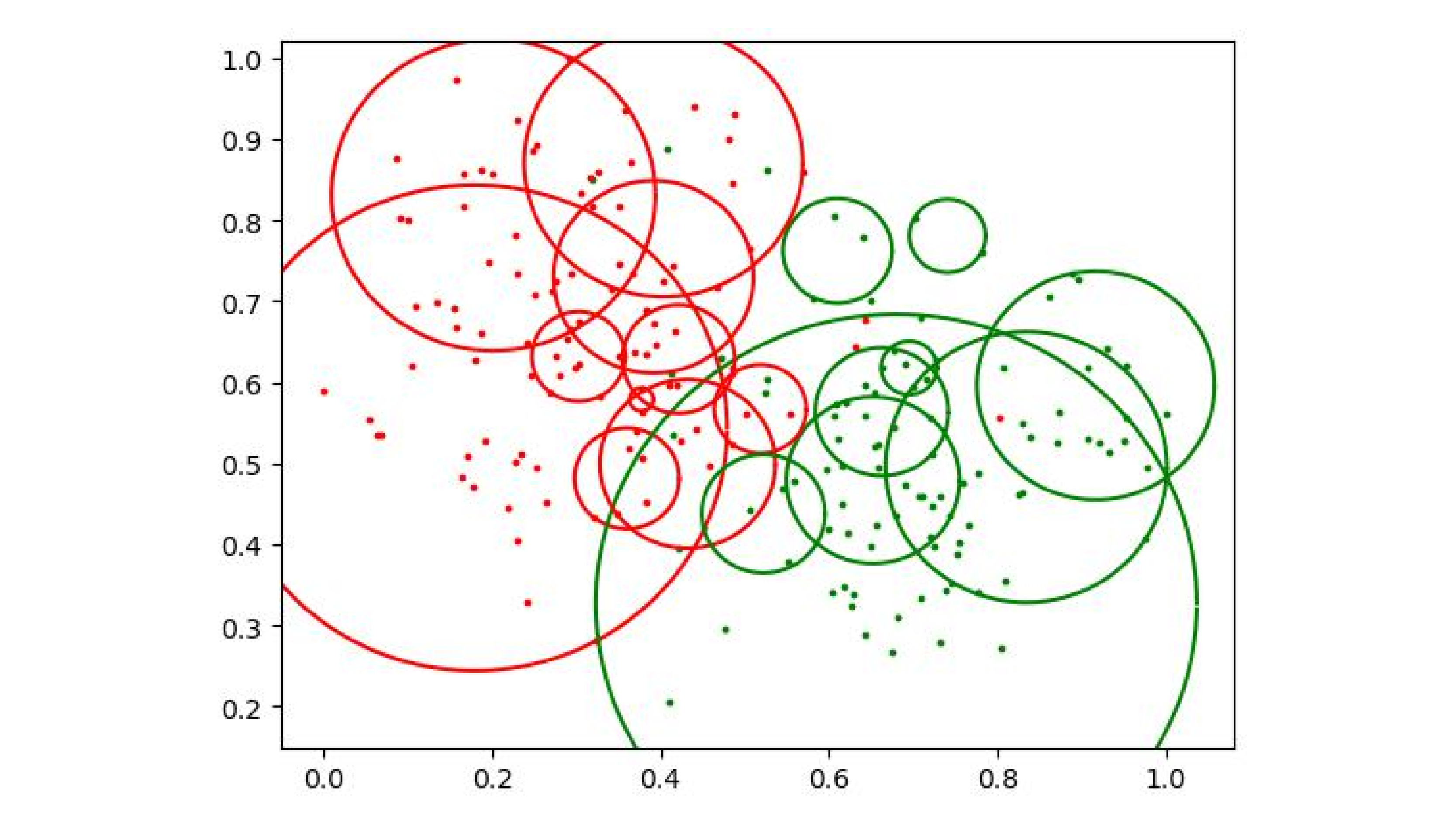}}
    	\subfigure[]{\includegraphics[width=0.24\textwidth]{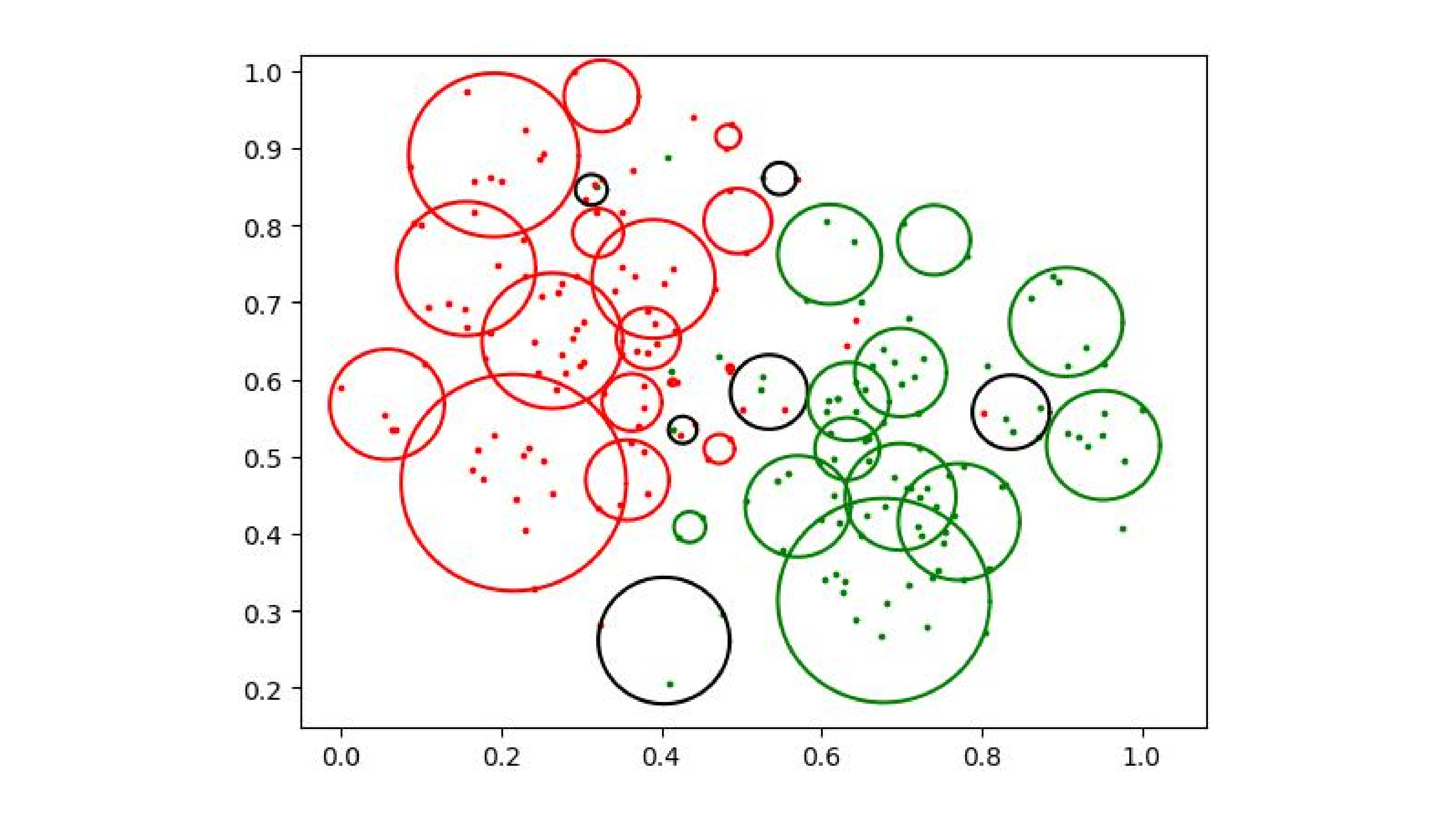}}
    	\caption{The generation process of GBNRS on the dataset. The red points and red granular-balls are labeled `+1',
    		and the green points and green granular-balls are labeled `-1'. The purity threshold \textit{$PT$} is 1 and \textit{$LBS$}'s value is 4. (a)-(e) The generation results of five iterations. (f) As a result of overlapping on the basis of (e), the purity of black granular-balls is less than 1, and the purity of red and green granular balls is 1.}
    	\label{fig:genbal} 
    \end{figure}

    Fig. \ref{fig:genbal} shows the granular-ball generation process of GBNRS. The red points and red granular-balls are labeled `+1', and the green points and green granular-balls are labeled `-1'. Firstly, the whole dataset is regarded as a granular-ball, it is divided into two granular-balls using 2-means because it contains two different classes of samples in it, and as shown in Fig. \ref{fig:genbal} (a). Fig. \ref{fig:genbal}(b), (c), (d) and (e) are the intermediate iteration results. For a granular-ball, if its purity $ < 1$ and its size $ \ge 4$, it continuous to be split. Fig. \ref{fig:genbal}(e) contains the phenomenon of heterogeneous transmission. It is eliminated by spliting the heterogeneous overlapped granular-balls to remove the overlap of heterogeneous balls, the results of which are shown in Fig. \ref{fig:genbal}(f). In Fig. \ref{fig:genbal}(f), any pair of two heterogeneous granular-balls do not contain any common samples. Besides, as shown in Fig. \ref{fig:genbal}(f), those granular-balls containing both green and red sample points, i.e., those black granualr-balls,  belong to the boundary region.

    \subsection{Granular-ball Rough Concept Tree for Knowledge Representation and Classification}
    
    \begin{figure}[hbpt!]
    	\centering
    	{\includegraphics[width = 0.5\textwidth]{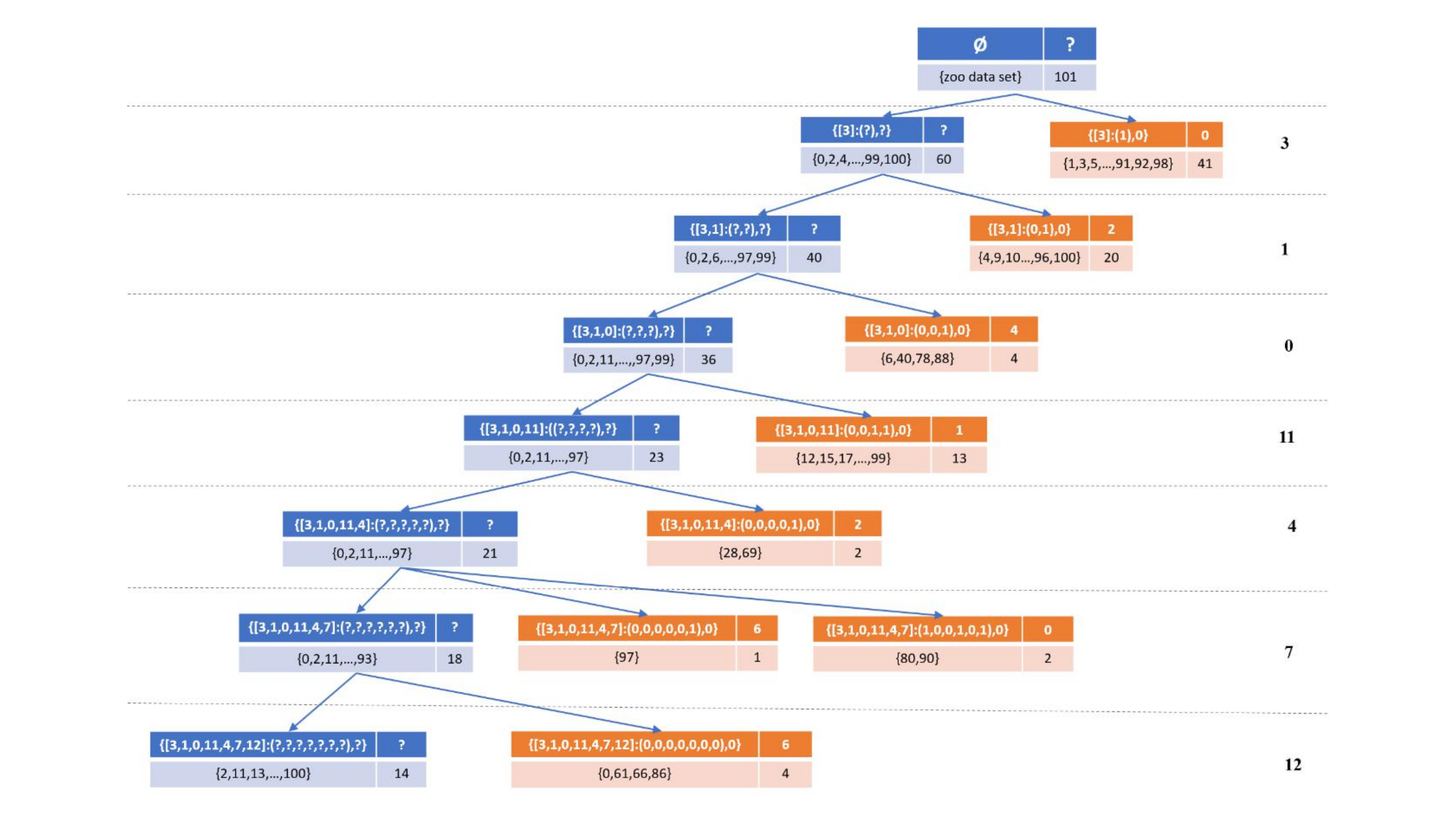}}
    	\caption{Knowledge representation of the discrete dataset zoo using GBRCT.}
    	\label{fig:zooRCT}
    \end{figure}
    \begin{figure*}[hbpt!]
    	\centering
    	{\includegraphics[width = 1.0\textwidth]{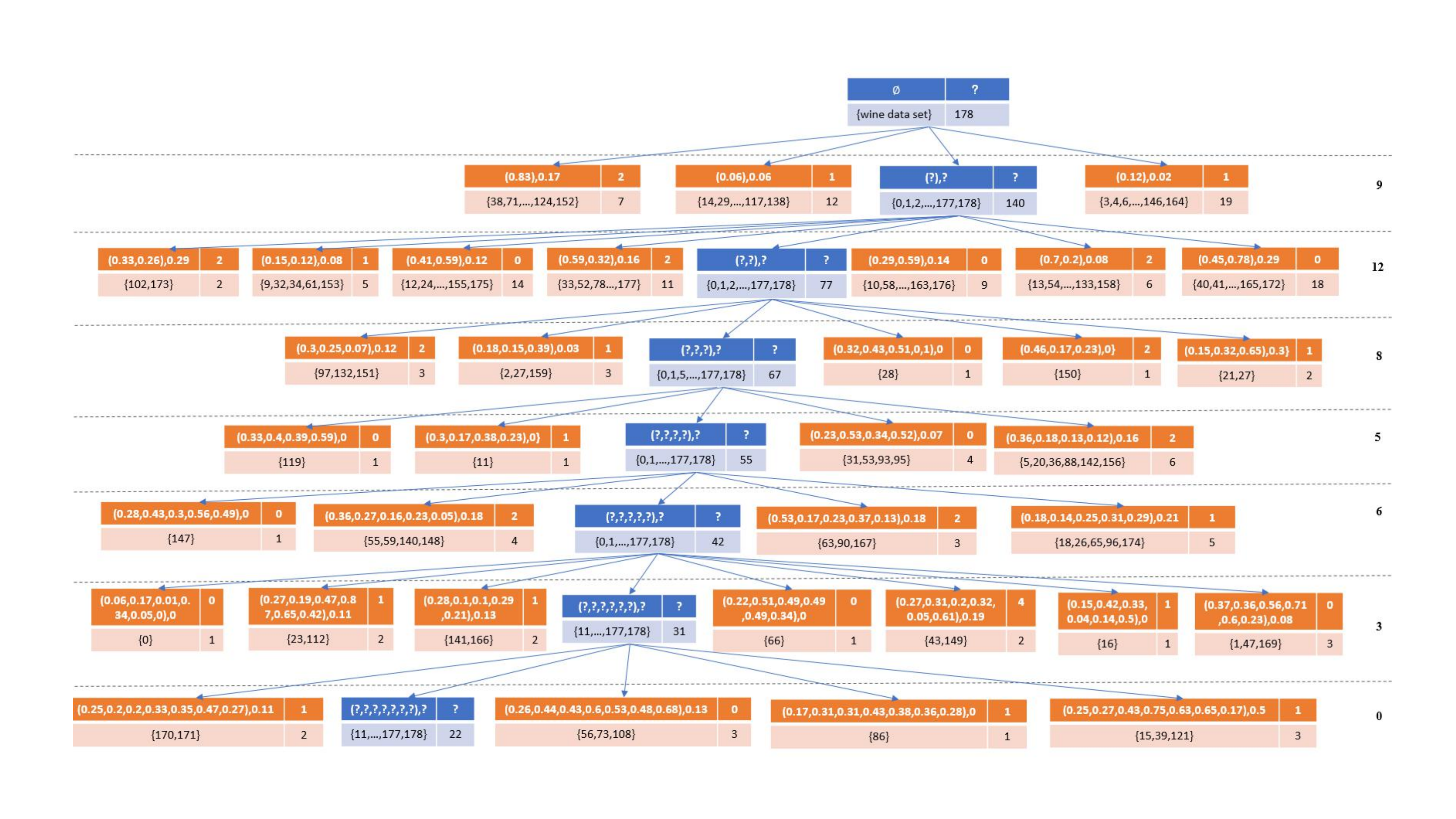}}
    	\caption{Knowledge representation of the continuous dataset wine using GBRCT.}
    	\label{fig:wineRCT}
    \end{figure*}
    
    As the GBRS can use equivalence class to represent upper and lower approximation while processing continuous data, it can represent knowledge well. In this section, we further proposed the granular-ball rough concept tree (GBRCT) by combing GBNRS with the rough concept tree \cite{xia2021rs}, which is proposed based on the concept lattice. RCT can not only be used to organize and describe the knowledge rules obtained by rough set based on forward attribute reduction algorithm, but also can be used for classification decision. So, the GBRCT makes GBRS a strong mining tool that not only can realize feature selection, but knowledge representation and classification at the same time.      
    
    In the RCT, each node is also called as a ``knowledge point'', or called ``concept node'', consisting of two parts: a sequence consisting of both some attributes and their values called ``intent'', and its corresponding equivalence classes called ``extent''. The ``concept" consisting with ``intent" and ``extent" is borrowed from philosophy for knowledge representation. GBRS strictly divides the dataset by gradular-balls, so in the GBRCT of this article, the representation of the first row and first column of the ``knowledge point'' has two parts: ``the attribute value'' and ``the center and radius of the corresponding gradular-ball''.     
    
    In the GBRCT, different from that in the RCT, the intent is described using a granular-ball equivalence class consisting of its center and radius instead of a sequence of attribute values. The GBRCT generated on the discrete dataset zoo is shown in Fig. \ref{fig:zooRCT}, and that on the continuous dataset wine is shown in Fig. \ref{fig:wineRCT}. Because we know the data set zoo is discrete in advance, that the neighberhood radius is smaller than a infinitely small positive value is used as the termination condition of granular-ball splitting, and GBRS is converted to be GBPRS. In Figs. \ref{fig:zooRCT} and \ref{fig:wineRCT}, an orange node represents a granular-ball equivalence class. A blue node who contains ``?'' represents boundary region, and an orange node representing a granular-ball belongs to the positive region that can certainly describe knowledge. The dataset zoo is discrete, so that the neighborhood radius is smaller than a infinitely small positive value is set to the termination condition of granular-ball splitting. The result of GBRCT is very similar with that using RCT except the represetation of the intent, i.e., the first row of each node. The RCT represents the intent using a sequence of attibute values, but the GBRCT using a granular-ball consisting of a center and a neighborhood radius whose value is equal to zero as shown in Fig. \ref{fig:zooRCT}; On the contrary, as shown in Fig. \ref{fig:wineRCT}, the neighborhood radius in the continious data is larger than zero. This also indicates that the GBRS realize the unified description for PRS and NRS well. In the real scene, when the information whether the data set is discrete or not is not provided in advance, an infinitely positive value is considered as an option to be optimized as the neighborhood radius in the GBRS.

    Similar with that in the RCT, the number in the right of a layer of the GBRCT shows which attributes the concept nodes in the layer are generated on; besides, those positive region concept nodes containing the largest number of extent samples have the strongest representation ability for knowledge and are the most valuable, such as the second node in the second layer and the second node in the third layer in Fig. \ref{fig:zooRCT}. In addition, as described in \cite{xia2021rs}, the GBRCT can also be directly used for classification.

\begin{algorithm}[htbp!]  
	\caption{Generation of granular-balls in the GBNRS}  
	\label{alg}
	\textbf{Input}: A dataset $ D=\{x_{1}, x_{2}, ..., x_{n}\} $, the lower bound of the size of the granular-ball $LBS$;  \\
	\textbf{Output}: $NOLGBs$; 
	\begin{algorithmic}[1]  		
		\STATE The current granular-ball set $CGBs$ and the granular-ball set in the next iteration $NGBs$ are initialized with $D$ and $\emptyset$;\\
		// Generate the initial granular-balls
		\REPEAT 
		\STATE $CGBs$ = $NGBs$; $NGBs$ = $\emptyset$; 
		\FOR{each granular-ball $GB_{i} \in CGBs $}
		\IF {$purity(GB_{i})<1$ and $|GB_{i}|>LBS$}
		\STATE Split $GB_{i}$ into $k$ sub-granular-balls $\{GB_{j}^{'},j=1,...,k\}$, where $k$ denotes the number of different labels in $GB_{i}$
		\STATE $NGBs$ = $NGBs$ + $\{GB_{j}^{'}\}$
		\ELSE 
		\STATE $NGBs$ = $NGBs$ + $GB_{i}$
		\ENDIF
		\ENDFOR
		\UNTIL {$|NGBs|==|CGBs|$}
		
		\STATE $OLGBs = \emptyset$; $NOLGBs = \emptyset$ \\
		//Remove the overlap between heterogeneous granular-balls
		\REPEAT	    	    
		\FOR{each granular-ball $GB_{i} \in CGBs $}
		\IF {there is overlap between $GB_{i}$ and $GB_{j} \in CGBs$ which has a different label}
		\STATE Split the larger granular-ball and add the sub-granular-balls into $OLGBs$
		\ELSE
		\STATE $NOLGBs$ = $NOLGBs$ + $GB_{i}$
		\ENDIF
		\ENDFOR
		\STATE $CGBs$ = $OLGBs$
		\UNTIL {$|OLGBs|=0$}		
	\end{algorithmic}  
\end{algorithm} 
       
    \subsection{Algorithm Design \label{sec:algori}}
    
    The only difference between the GBNRS and PRS is the generation way of positive region, and the feature selection process is the same because the GBNRS and PRS have an unified representation model. So, we only discuss the algorithm design of granular-ball generation for positive region in this section, which is shown in Algorithm 1. The algorithm 1 mainly consists of two parts including initial granular-balls generation and overlap removing. $purity(GB)$ denotes the purity of the granular-ball $GB$. In Step 16, there is overlap between two granular-balls if their boudary distance is smaller than zero, i.e,that, the distance between their centers is smaller than sum of their radii. The process of spliting in Step 17 is the similar with that in Step 6. In the output variable $NOLGBs$, those granular-balls with purity as 1 belong to positive region. The lower bound of the size of a granular ball $LBS$, i.e., the number of samples in it, is optimized from 2*$d$ to 2 with a step as 1, where the $d$ denotes the number of conditional attributes in the data set.

    \section{Experiment}\label{sec:experiment}
    
    To demonstrate the feasibility and effectiveness of GBRS, we selected some popular or the state-of-the-art algorithms for comparison. As the experimental results using GBPRS are the same with those using PRS, so the GBNRS is selected for comparison. As the PRS can only process discrete data, so we also need to conduct experiments on some discrete data sets for comparison with PRS. As shown in Table \ref{tab:1}, we randomly selected fifteen real datasets including continuous and discrete datasets to demonstrate the performance of the GBNRS. Among them, the first six are discrete datasets; the last nine are continuous datasets. Experimental hardware environment: PC with an Intel Core i7-107000 CPU @2.90 GHz with 32 G RAM. Experimental software environment: Python 3.7.

    \begin{table}[!ht]
    	\centering
    	\caption{Dataset Information}
    	\label{tab:1}
    	\setlength{\tabcolsep}{1.2mm}{
    		\begin{tabular}{clcccc}
    			
    			\toprule
    			
    			\multirow{3}{*}{\begin{tabular}[c]{@{}l@{}}NO.\end{tabular}} & \multirow{3}{*}{\begin{tabular}[c]{@{}l@{}}Dataset\end{tabular}} & \multirow{3}{*}{\begin{tabular}[c]{@{}l@{}}Samples\end{tabular}} & 
    			\multirow{3}{*}{\begin{tabular}[c]{@{}l@{}}Numerical \\ Condition \\Attributes \end{tabular}}&
    			\multirow{3}{*}{\begin{tabular}[c]{@{}l@{}}Categorical \\ Condition \\Attributes \end{tabular}}&
    			\multirow{3}{*}{\begin{tabular}[c]{@{}l@{}}Class\end{tabular}}	
    			\\ \\
    			&                                                                         &                           &    &                                                                                 \\
    			\midrule
    			1&	lymphography&	148&	0&	18&	4\\
    			2&	primary-tumor&	336&	0&	15&	2\\
    			3&	mushroom&	7535&	0&	22&	2\\
    			4&	mushroom1&	8124&	0&	22&	2\\	
    			5&	zoo&	101&	0&	16&	7\\
    			6&	backup-large&	307&	0&	36&	4\\
    			7&	iono&	351&	34&	0&	2\\
    			8&	Diabetes&	768&	8&	0&	2\\
    			9&	wdbc&	569&	30&	0&	2\\
    			10&	audit\_risk	&772&	21&	0&	2\\
    			11&	electrical	&10000&	13&	0&	2\\ 
    			\multirow{2}{*}{12} & Parkinson\_Multiple  & \multirow{2}{*}{1040} & \multirow{2}{*}{27} & \multirow{2}{*}{0}& \multirow{2}{*}{2}\\
    			& \_Sound\_Recording & & & & \\
    			13&	wine	&178	&13&	0&	3\\
    			14&	spambase&	4601&	57&	0&	2\\
    			15&	htru2&	17898&	8&	0&	2\\
    			
    			\bottomrule
    		\end{tabular}
    	}
    \end{table}
    
    The lower bound of the size of a granular ball $LBS$, i.e., the number of samples in it, is optimized from 2*$d$ to 2 with a step as 1, where the $d$ denotes the number of conditional attributes in the data set. The experiments are designed along the lines of Xia et al. \cite{54} as the quality of the reduced attribute set is not related to the testing classifier used, only a common testing classifier the nearest neighbor algorithm—is used to verify the quality of the reduced attribute set. Therefore, we use the common classifier, kNN, in our experiment with 5-fold cross-validation. 
    
    \subsection{In Comparison with PRS under Discrete Data} \label{section 4.1.1}
     The experimental results on the first six discrete datasets in Table \ref{tab:1} are shown in Table \ref{tab:2}, where the "original" column represents the classification accuracy obtained from the original unreduced dataset. The "NO" column in Table \ref{tab:2} is corresponded with the "NO" column in Table \ref{tab:1}. It can be seen from Table \ref{tab:2} that, the classification accuracy of GBNRS is much higher than that of the PNS on most cases except the case on the 4$^{th}$ data set, in which the accracies of the two algorithms are the same. Considering the average classification accuracy, the original accuracy and PRS's accuray are 0.8906 and 0.8618 respectively, while that of GBNRS is 0.8958; in comparison with the previous two results, GBNRS achieved 0.52 and 3.4 percentage enhancement respectively. The reason is that, the GBNRS can flexibly fit the data distribution using those granular-balls with varies radii, which is obviously better than those methods using a fixed radius, such as PRS and the traditional NRS. So, GBNRS can achieve a higher accuracy than the two algorithms. In summary, analysis shows that on discrete datasets, GBNRS can achieve higher classification accuracy than both the PRS and those on the original datasets.

    \begin{table}[!htbp]
    	\centering
    	
    	\caption{Accuracy Comparison on Discrete Datasets}
    	\label{tab:2}
    	\setlength{\tabcolsep}{3mm}{
    		\begin{tabular}{cccc}
    			\toprule
    			NO.     & Original      & PRS        & GBNRS  \\
    			\midrule
    			1       & 0.8101$\pm$0.0966 & 0.7489$\pm$0.0627 & \textbf{0.8161$\pm$0.0584} \\
    			2       & \textbf{0.6955$\pm$0.0290} & 0.6686$\pm$0.0414 & \textbf{0.6955$\pm$0.0290} \\
    			3       & 0.9111$\pm$0.1034 & 0.9202$\pm$0.0753 & \textbf{0.9328$\pm$0.0746} \\
    			4       & \textbf{1$\pm$0}   & 0.9889$\pm$0.0026 & \textbf{1$\pm$0}           \\
    			5       & 0.95$\pm$0.0499 & 0.9$\pm$0.079 & \textbf{0.95$\pm$0.0353} \\
    			6       & 0.9771$\pm$0.0274 & 0.9444$\pm$0.0526 & \textbf{0.9804$\pm$0.0269} \\
    			Average & 0.8906        & 0.8618        & \textbf{0.8958}   \\      
    			
    			\bottomrule
    		\end{tabular}
    	}
    \end{table}
         
    \subsection{In Comparsion with Varies Feature Selection Methods} \label{section 4.1.2}
    in this section, we select nine continuous datasets whose indexes are from 7 to 15 in Table \ref{tab:1}, and nine popular or the-state-of-the-art algorithms for comparison including NRS \cite{63}, GBNRS$_{old}$ \cite{54}, Cfs \cite{32}, Ilfs \cite{01}, Laplacian \cite{60}, Lasso \cite{61}, Mrmr \cite{68}, WNRS \cite{64}. The experimental results are shown in Table \ref{tab:3}. The neighborhood radius $\delta$ is gradually increased from 0.01 to 0.5 with a step size of 0.01, which is commonly used in the NRS and WNRS. The method, GBNRS$_{old}$, also introduce granular-ball computing to decrease the overlap in the traditional NRS, resulting in the efficiency improvement. However, it did not realize the eqivalence representation; so, we named it with "old" as suffix, i.e., GBNRS$_{old}$, for distinguishing it from our method GBNRS. As there is randomness in the GBNRS$_{old}$, we run GBNRS ten times, and take the highest classification accuracy among the ten experiments results for comparison. The experimental results are shown in Table \ref{tab:3}. It can be seen from Table \ref{tab:3} that, in comparison with other algorithms, GBNRS can achieve the highest classification accuracy in seven datasets on most cases. The reason is that the combination of the robustness and adaptability of the granular-ball computing. The adaptability makes GBNRS flexibly fit different data distributions using those granular-balls with varies radii, resulting in a good performance of GBNRS in accuracy.

    \begin{table*}[htbp!]
    	
    	\centering
    	
    	\caption{Accuracy of different attribute reduction algorithms on continuous datasets}
    	\label{tab:3}
    	\setlength{\tabcolsep}{0.4mm}{
    		\begin{tabular}{ccccccccccc}
    			\toprule
    			NO.  & Cfs & Ilfs &Laplacian &Lasso &Mrmr &Original &NRS &GBNRS$_{old}$ &WNRS &GBNRS          \\
    			\midrule
    			1    &0.8234$\pm $0.0589&0.8029$\pm $0.0551&0.7963$\pm $0.1125 &0.7687$\pm $0.0441
    			&\textbf{0.8239$\pm $0.0715}&0.8101$\pm $0.0966&0.7566$\pm $0.1094 &0.8035$\pm $0.0575 &0.7489$\pm$0.0627 &0.8161$\pm $0.0584\\
    			
    			2    &0.7045$\pm $0.0591&\textbf{0.7224$\pm $0.0467}&0.6626$\pm $0.0479 &0.7105$\pm $0.0467
    			&0.6388$\pm $0.0771&0.6955$\pm $0.0209&0.6686$\pm $0.0414 &0.6716$\pm $0.0606 &0.6687$\pm$0.0414 &0.6955$\pm $0.0290\\
    			
    			3    &0.8997$\pm $0.1235&0.9040$\pm $0.1420&0.9146$\pm $0.1080 &0.9247$\pm $0.1029
    			&0.9588$\pm $0.0677&0.9111$\pm $0.1035&0.9729$\pm $0.0260 &0.9289$\pm $0.0755 &\textbf{0.9745$\pm$0.0571} &0.9328$\pm $0.0746\\
    			
    			4    &\textbf{1$\pm $0}&\textbf{1$\pm $0}&\textbf{1$\pm $0} &\textbf{1$\pm $0}
    			&\textbf{1$\pm $0}&\textbf{1$\pm $0}&\textbf{1$\pm $0} &\textbf{1$\pm $0} &\textbf{1$\pm$0} &\textbf{1$\pm $0}\\
    			
    			5    &0.8900$\pm $0.0418&0.9300$\pm $0.0273&0.9200$\pm $0.0274 &0.9200$\pm $0.0570
    			&\textbf{0.9600$\pm $0.0547}&0.9500$\pm $0.0499&0.9000$\pm $0.079 &0.9100$\pm $0.0418 &0.9000$\pm$0.0790 &0.9500$\pm $0.0353\\
    			6    &0.9771$\pm $0.0274&\textbf{0.9804$\pm $0.0269}&0.9771$\pm $0.0274 &0.9738$\pm $0.034
    			&0.9771$\pm $0.0274&0.9771$\pm $0.0274&0.9705$\pm $0.0408 &0.964$\pm $0.0549 &0.9771$\pm$0.0249 &\textbf{0.9804$\pm $0.0269}\\
    			7    &0.9028$\pm $0.0340&0.8800$\pm $0.0372&0.8514$\pm $0.0312 &0.8657$\pm $0.0423
    			&0.9142$\pm $0.0319&0.8486$\pm $0.0458&0.8886$\pm $0.0275 &0.8771$\pm $0.0186 &\textbf{0.9146$\pm$0.0211} &0.9000$\pm $0.0484\\
    			
    			8   &0.6897$\pm $0.0136&0.7431$\pm $0.0250&0.7327$\pm $0.0315 &0.6897$\pm $0.0412
    			&0.7210$\pm $0.0463&\textbf{0.7471$\pm $0.0348}&\textbf{0.7471$\pm $0.0348} &\textbf{0.7471$\pm $0.0348}&\textbf{0.7471$\pm $0.0348}&\textbf{0.7471$\pm $0.0348}\\
    			
    			9   &0.9666$\pm $0.0273&0.9701$\pm $0.0192&0.9718$\pm $0.0266 &0.9718$\pm $0.0266&0.9718$\pm $0.0259&0.9683$\pm $0.0253&0.9701$\pm $0.0192 &0.9665$\pm $0.0243 &0.9613$\pm$0.022&\textbf{0.9718$\pm $0.0218}\\
    			
    			10   &0.9377$\pm $0.6070&0.9468$\pm $0.0360&0.9559$\pm $0.0451 &0.9624$\pm $0.0213&0.9688$\pm $0.0202&0.9377$\pm $0.0571&0.9195$\pm $0.0698 &0.9416$\pm $0.5728 &0.9235$\pm$0.0718&\textbf{0.9922$\pm $0.0071}\\

    			11   &0.8594$\pm $0.0081&0.9721$\pm $0.0020&0.8594$\pm $0.0081 &0.8594$\pm $0.0081&0.9147$\pm $0.0068&0.9145$\pm $0.0068&0.9721$\pm $0.0058 &0.9145$\pm $0.0068 &0.9782$\pm$0.0022&\textbf{0.9969$\pm $0.0015}\\
    			
    			12    &0.9057$\pm $0.0455&0.8249$\pm $0.0686&0.8913$\pm $0.0522 &0.8970$\pm $0.0512&0.9259$\pm $0.5930&0.8249$\pm $0.6860&\textbf{1$\pm $0} &0.8797$\pm $0.0053 &\textbf{1$\pm $0} &\textbf{1$\pm $0} \\
    			
    			13    &0.9435$\pm $0.0282&0.9662$\pm $0.0235&0.9719$\pm $0.0340 &0.9660$\pm $0.0310&0.9775$\pm $0.0233&0.9605$\pm $0.0427&\textbf{0.9830$\pm $0.0155}&0.9438$\pm $0.0478&0.9721$\pm$0.0196&0.9773$\pm $0.0238 \\
    			
    			14    &0.8715$\pm $0.0562&0.8609$\pm $0.0582&0.8661$\pm $0.0559 &0.8806$\pm $0.2870&0.8830$\pm $0.0603&0.8646$\pm $0.0639&0.8648$\pm $0.0650&0.8689$\pm $0.0634&0.8646$\pm$0.0646 &\textbf{0.8889$\pm $0.0531} \\
    			
    			15    &0.9772$\pm $0.0021&0.9772$\pm $0.0021&0.9772$\pm $0.0021 &0.9777$\pm $0.0015&0.9772$\pm $0.0021&0.9772$\pm $0.0021&0.9772$\pm $0.0021&0.9772$\pm $0.0021 &0.9778$\pm $0.0021&\textbf{0.9785$\pm $0.0021} \\
    			\bottomrule
    		\end{tabular}
    	}
    \end{table*}

    \section{Conclusion}\label{sec:conclusion}
    This paper presents an unified model for the two most popular rough set models, Pawlak rough set and neighborhood rough set models. The unified model can not only express knowledge with equivalence classes, but also deal with both continuous data and discrete data. In comparison with nine popular or the-state-of-the-art feature selection methods on fifteen real datasets, the experiments show that the proposed model can achieve a better performance in accuracy. However, the optimization of the $LBS$ is inefficient. If some incremental strategies can be developed, it can be accelerated. In the future work, we will developed the poposed model into other rough set models and improve their performance.

	\section{Acknowlegements}
	This work was supported in part by the National Natural Science Foundation of China under Grant Nos. 62176033 and 61936001, the Natural Science Foundation of Chongqing under Grant No. cstc2019jcyj-cxttX0002 and by NICE: NRT for Integrated Computational Entomology, US NSF award 1631776, the National Key Research and Development Program of China under Grant No. 2019QY(Y)0301.


\begin{IEEEbiography}[{\includegraphics[width=1in,height=1.25in,clip,keepaspectratio]{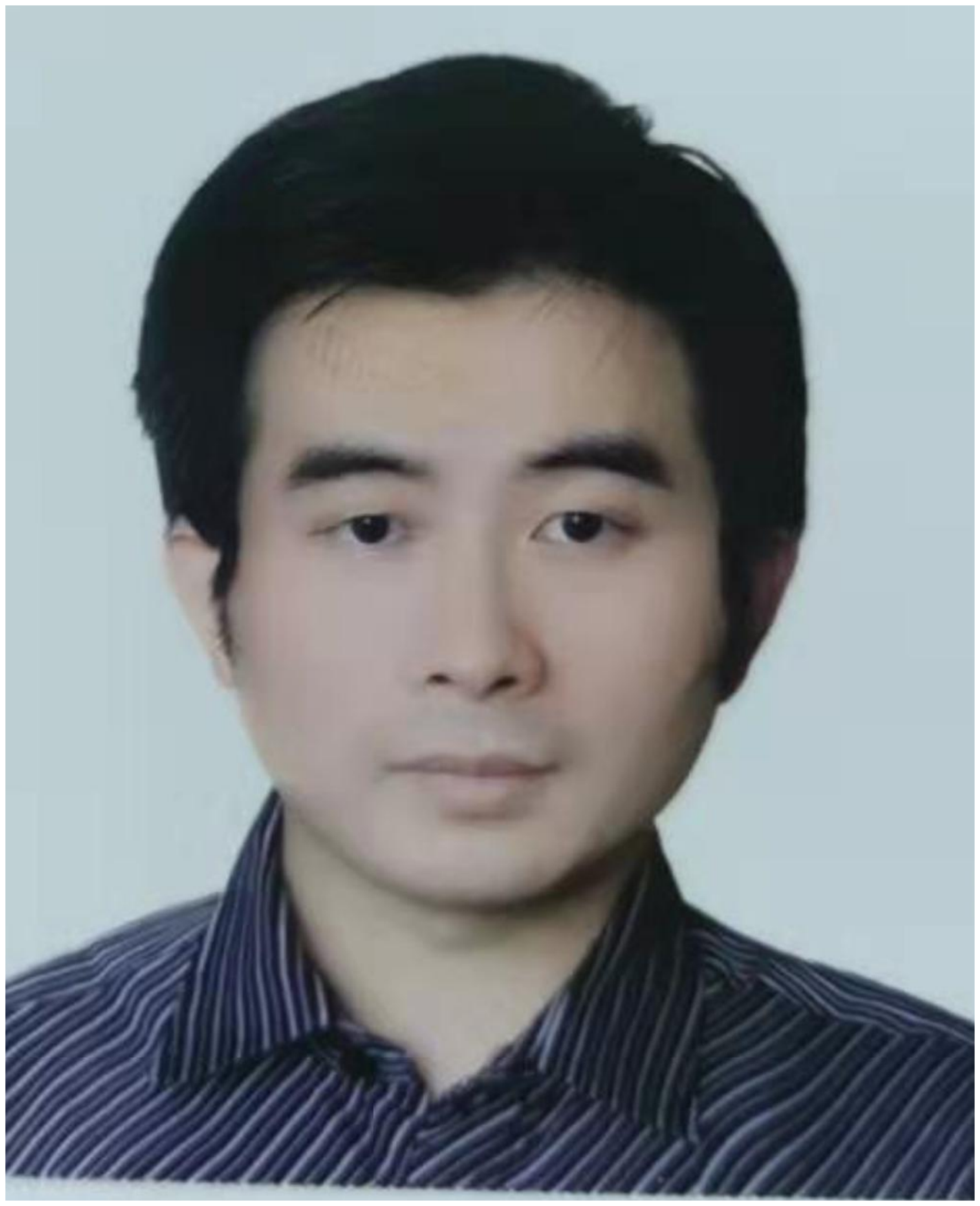}}]{Shuyin Xia} received his B.S. and M.S. degrees in computer science from the Chongqing University of Technology in China in 2008 and 2012. He received his Ph.D. degree from the College of Computer Science at Chongqing University in China. He is an associate professor and a Ph.D. supervisor at the Chongqing University of Posts and Telecommunications in Chongqing, China. His research interests include classifiers and granular computing. He has published more than 30 papers in international journals and conferences, such as IEEE T-PAMI, IEEE T-KDE, T-NNLS, T-CYB and IS. He is the executive deputy director of the Big Data and Network Security Joint Lab of CQUPT and an IEEE Member.
\end{IEEEbiography}

\vspace{-3em}
\begin{IEEEbiography}[{\includegraphics[width=1in,height=1.25in,clip,keepaspectratio]{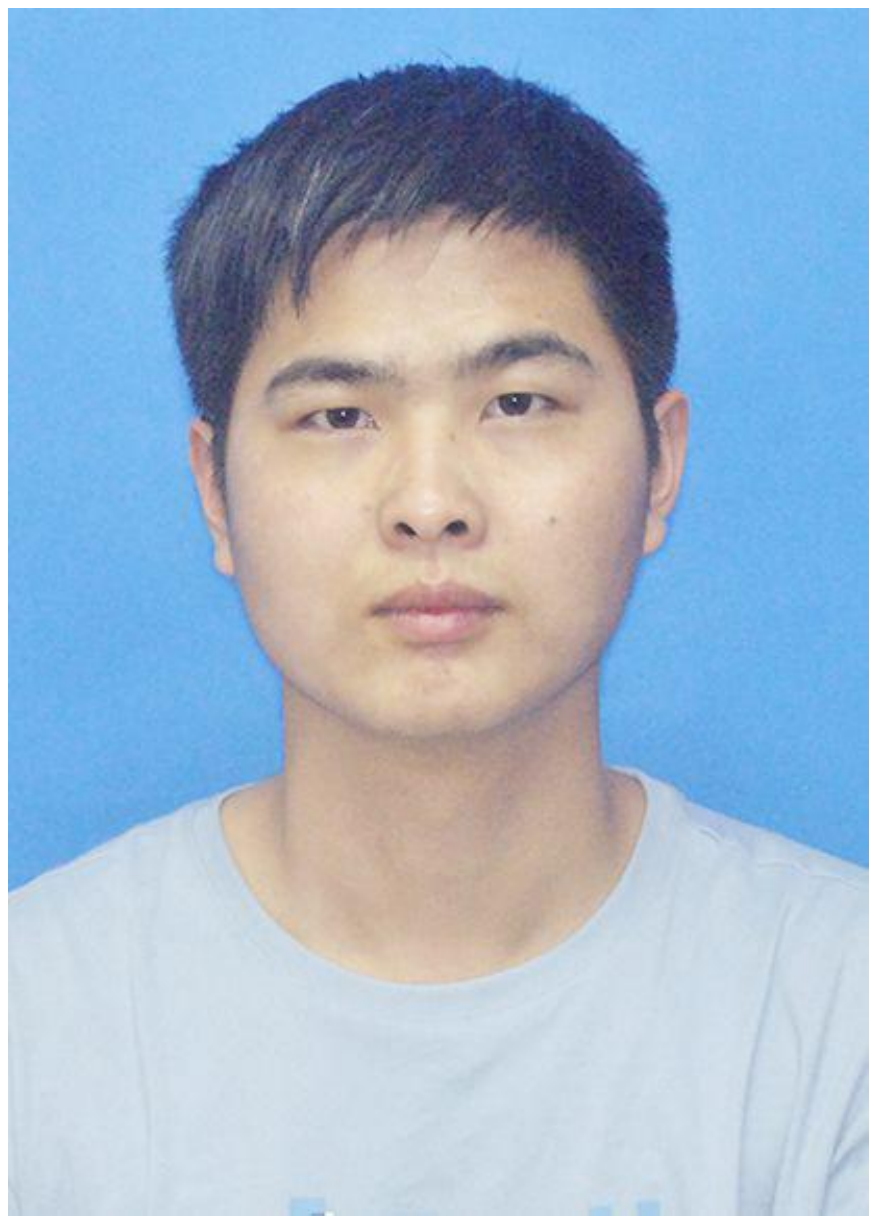}}]{Cheng Wang} received his B.S. degree in information and computer science from the ChangAn University in China. He is currently a graduate student in the College of Computer Science and Technology at the Chongqing University of Posts and Telecommunications. His research interests include machine learning and rough set theory.
\end{IEEEbiography}

\vspace{-3em}
\begin{IEEEbiography}[{\includegraphics[width=1in,height=1.25in,clip,keepaspectratio]{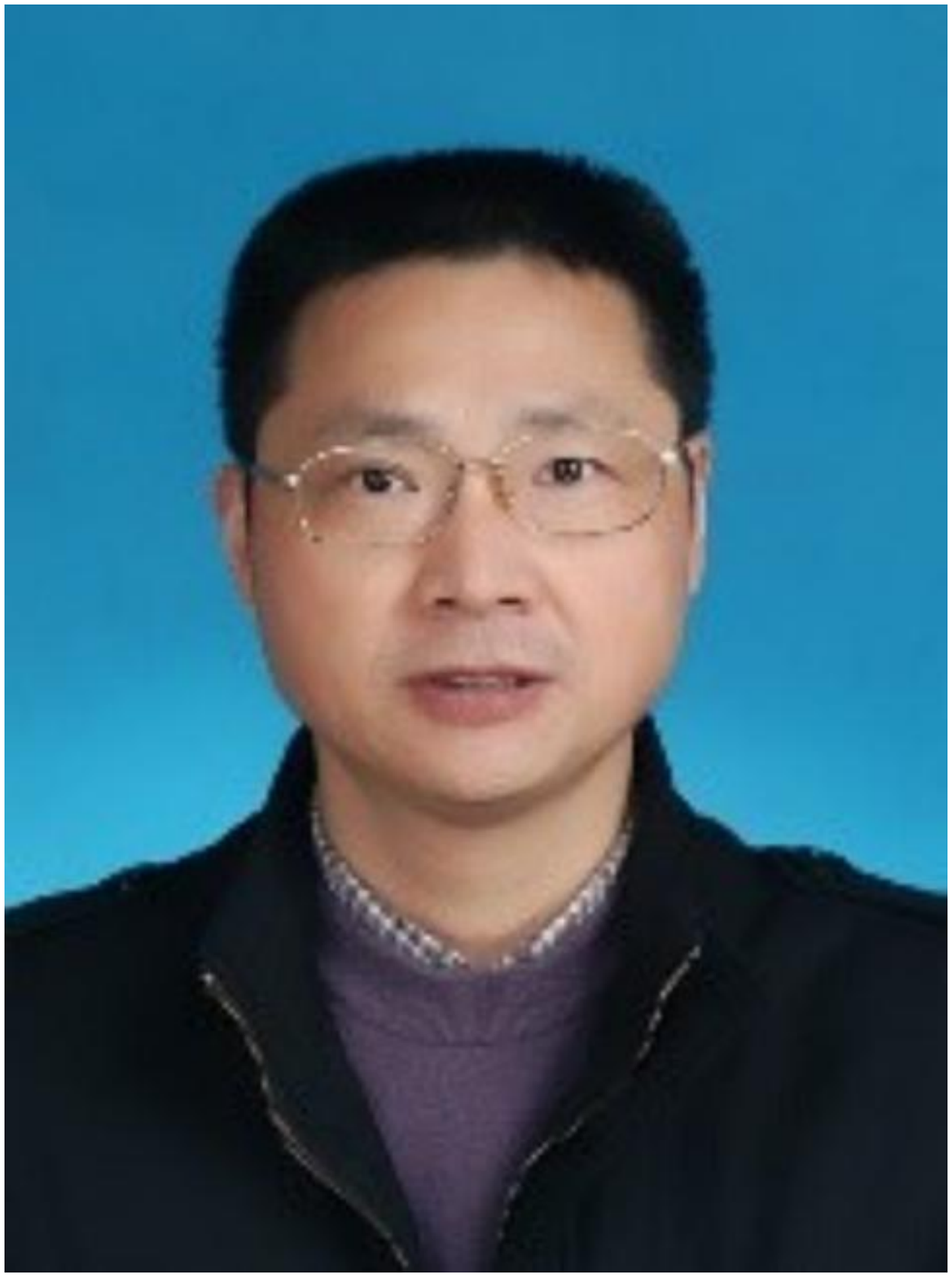}}]{Guoyin Wang} received B.E and M.S. degrees in computer software from Xi'an Jiaotong University in Xi'an, China in 1992 and 1994. He received his Ph.D. degree in computer organization and architecture in 1996 from Xi'an Jiaotong University. His research interests include data mining, machine learning, rough sets, granular computing, cognitive computing, and so forth. He has published over 300 papers in international journals and conferences, including IEEE T-PAMI, T-KDE, T-IP, T-NNLS, and T-CYB. He has worked at the University of North Texas, USA, and the University of Regina, Canada, as a Visiting Scholar. Since 1996, he has been working at the Chongqing University of Posts and Telecommunications in Chongqing, China, where he is currently a Professor and a Ph.D. supervisor, the Director of the Chongqing Key Laboratory of Computational Intelligence, and the Vice President of the Chongqing University of Posts and Telecommunications. He is the Steering Committee Chair of the International Rough Set Society (IRSS), a Vice-President of the Chinese Association for Artificial Intelligence (CAAI), and a council member of the China Computer Federation (CCF).\end{IEEEbiography}

\vspace{-3em}
\begin{IEEEbiography}[{\includegraphics[width=1in,height=1.25in,clip,keepaspectratio]{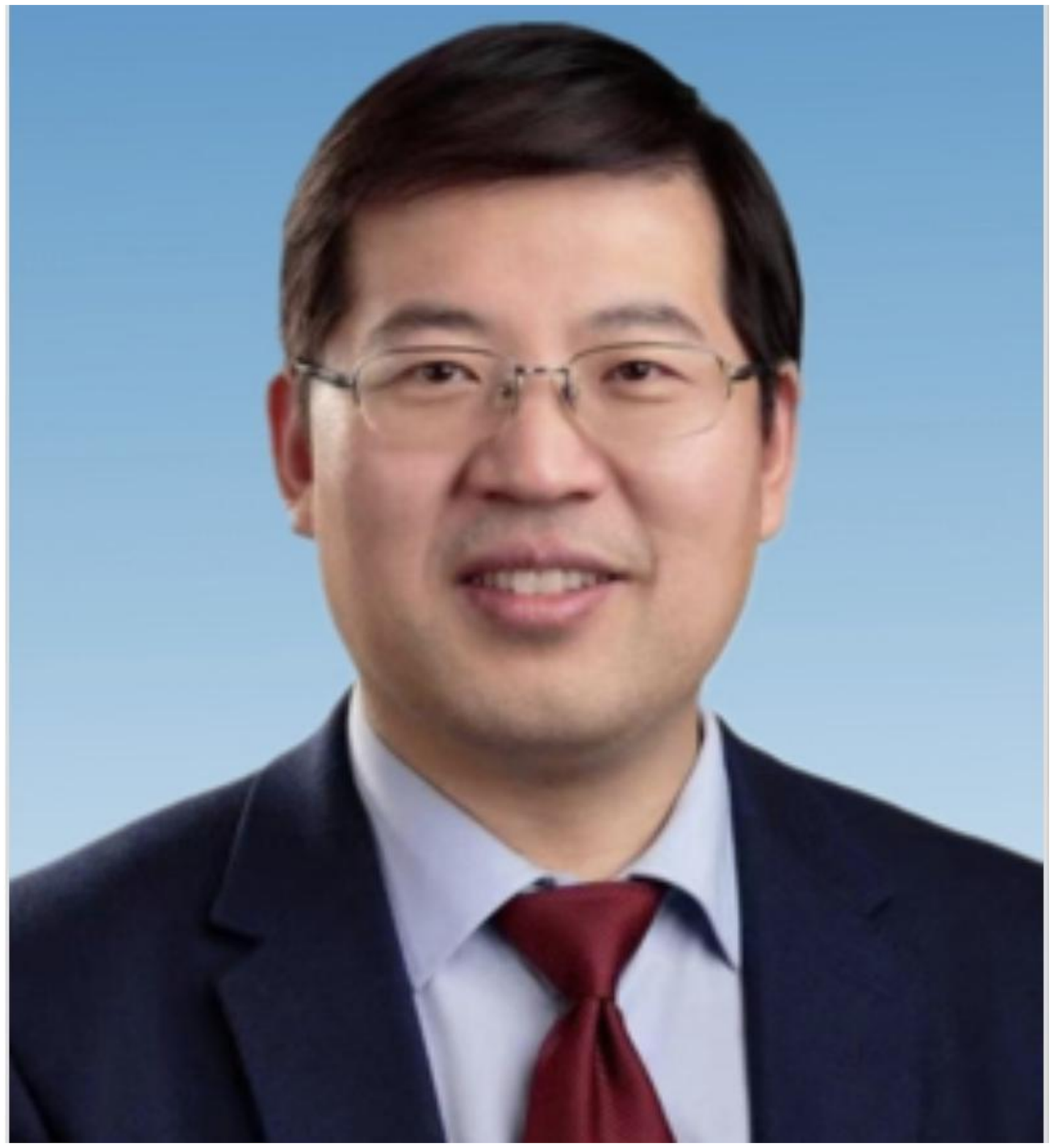}}]{Xinbo Gao} (M’02-SM’07) received BEng, MSc, and PhD degrees in signal and information processing from Xidian University in Xi’an, China, in 1994, 1997, and 1999. From 1997 to 1998, he was a research fellow with the Department of Computer Science at Shizuoka University in Shizuoka, Japan. From 2000 to 2001, he was a post-doctoral research fellow with the Department of Information Engineering at the Chinese University of Hong Kong in Hong Kong. From 2001 to 2020, he was with the School of Electronic Engineering at Xidian University. He is currently the President of the Chongqing University of Posts and Telecommunications. He has published six books and over 200 technical articles in international journals and conferences, including IEEE T-PAMI, T-IP, T-NNLS, T-MI, NIPS, CVPR, ICCV, AAAI and IJCAI.    
\end{IEEEbiography}

\begin{IEEEbiography}[{\includegraphics[width=1in,height=1.25in,clip,keepaspectratio]{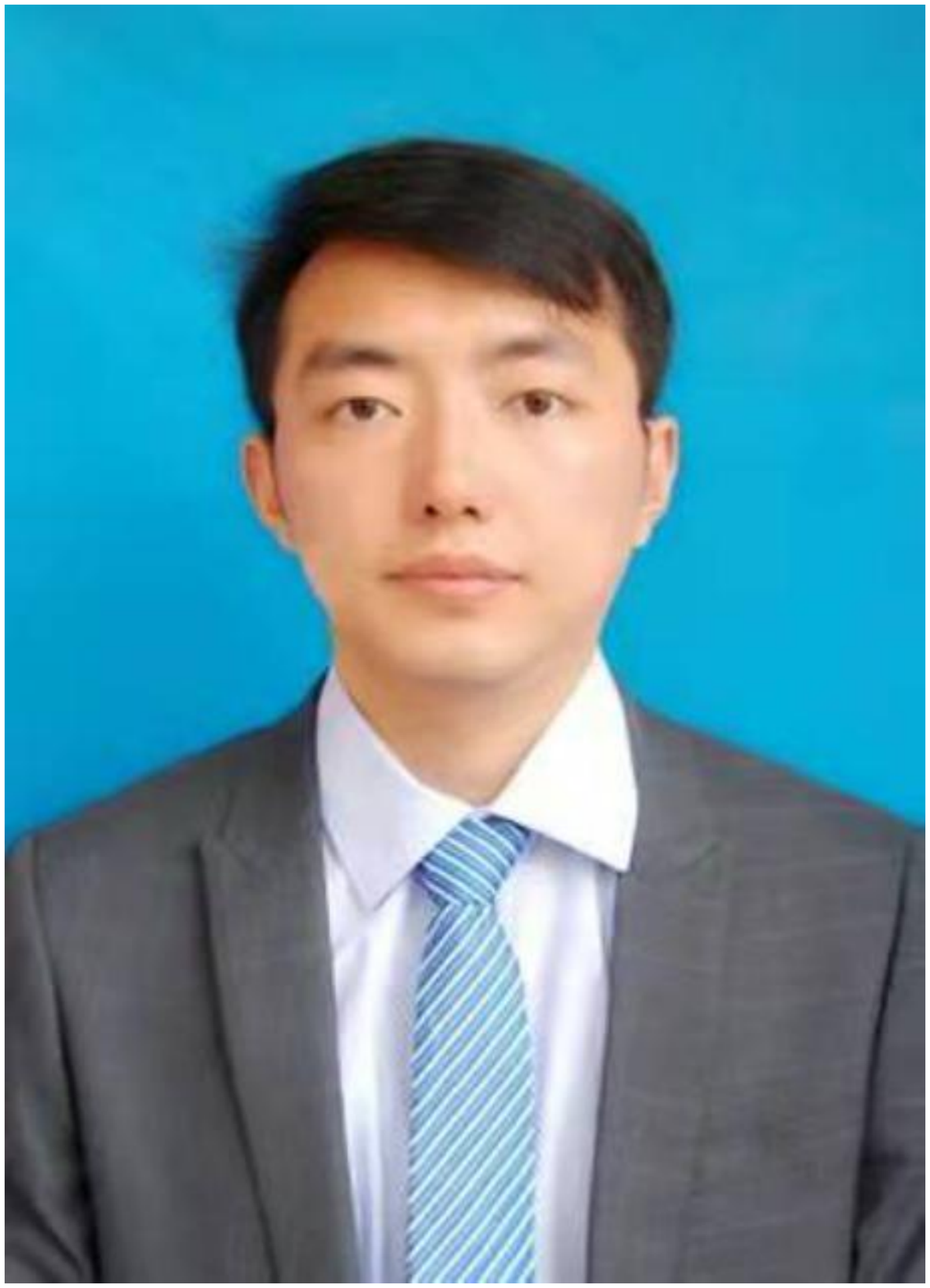}}]{Jianhang Yu} received his B.S. and M.S. degrees in information and computing science from the Chongqing University of Technology, China, in 2013 and 2016. He received his Ph.D. degree from the School of Mathematics at Harbin Institute of Technology in China. From 2018 to 2019, he was a visiting scholar with the Graduate School of Information Science at Osaka University in Suita, Japan. His research interests include fuzzy sets, rough sets, decision-making analysis, granular computing and cognitive computing.
\end{IEEEbiography}

\end{document}